\documentclass[12pt]{article}
\usepackage{framed,epsf,latexsym,amsmath,
amssymb,amscd,mathrsfs}
\usepackage[round,comma]{natbib}                                 
\usepackage{graphicx}
\usepackage[pdftex,colorlinks]{hyperref}
\usepackage[margin=1.0in]{geometry}

\newtheorem{theorem}{Theorem}[section]
\newtheorem{lemma}[theorem]{Lemma}
\newtheorem{remark}[theorem]{Remark}

\newtheorem{counter-example}[theorem]{Counter example}
\newtheorem{proposition}[theorem]{Proposition}
\newtheorem{open question}[theorem]{Open question}
\newtheorem{corollary}[theorem]{Corollary}

\newtheorem{definition}[theorem]{Definition}

\newtheorem{claim}{Claim}

\newcommand{\ignore}[1]{}

\newcommand{\ca}{{\cal A}}

\newcommand{\cd}{{\cal D}}

\newcommand{\cm}{{\cal M}}

\newcommand{\cf}{{\cal F}}

\newcommand{\cp}{{\cal P}}

\newcommand{\cx}{{\cal X}}
\newcommand{\cy}{{\cal Y}}
\newcommand{\cz}{{\cal Z}}

\DeclareMathOperator*{\hinge}{hinge}
\DeclareMathOperator*{\sign}{sign}

\newcommand{\reals}{{\mathbb R}}

\newcommand{\complex}{{\mathbb C}}

\newcommand{\proof}{{\par\noindent {\bf Proof}\space\space}}
\newcommand{\proofbox}{\begin{flushright}$\Box$\end{flushright}}

\DeclareMathOperator{\Err}{Err}
\DeclareMathOperator{\poly}{poly}
\DeclareMathOperator*{\E}{\mathbb{E}}

\newcommand{\D}{\mathcal{D}}
\newcommand{\inner}[1]{\langle #1 \rangle}

\title{
The complexity of learning halfspaces using generalized linear methods
}
\author{Amit Daniely \thanks{Department of Mathematics, Hebrew
                 University, Jerusalem 91904, Israel.  amit.daniely@mail.huji.ac.il}
                 \and Nati Linial \thanks{School of Computer Science and Engineering, Hebrew University, Jerusalem 91904, Israel.  nati@cs.huji.ac.il}
                 \and Shai Shalev-Shwartz\thanks{School of Computer Science and Engineering, Hebrew University, Jerusalem 91904, Israel. shais@cs.huji.ac.il}}

\begin{document}

\maketitle
\setcounter{page}{0}

\thispagestyle{empty}

\maketitle

\begin{abstract}
Many popular learning algorithms (E.g. Regression, Fourier-Transform based algorithms, Kernel SVM and Kernel ridge regression) operate by reducing the problem to a convex optimization problem over a set of functions. These methods offer the currently best approach to several central problems such as learning half spaces and learning DNF's. In addition they are widely used in numerous application domains. Despite their importance, there are still very few proof techniques to show limits on the power of these algorithms.

We study the performance of this approach in the problem of
(agnostically and improperly) learning halfspaces with margin
$\gamma$. Let $\D$ be a distribution over labeled examples. The
$\gamma$-margin error of a hyperplane $h$ is the probability of an
example to fall on the wrong side of $h$ or at a distance $\le\gamma$
from it. The $\gamma$-margin error of the best $h$ is denoted
$\Err_\gamma(\D)$.  An $\alpha(\gamma)$-approximation algorithm
receives $\gamma,\epsilon$ as input and, using i.i.d. samples of $\D$,
outputs a classifier with error rate $\le
\alpha(\gamma)\Err_\gamma(\D) + \epsilon$.  Such an algorithm is
efficient if it uses $\poly(\frac{1}{\gamma},\frac{1}{\epsilon})$
samples and runs in time polynomial in the sample size.

The best approximation ratio achievable by an efficient algorithm is
$O\left(\frac{1/\gamma}{\sqrt{\log(1/\gamma)}}\right)$ and is achieved
using an algorithm from the above class. Our main result shows that the
approximation ratio of every efficient algorithm from this family must be $\ge
\Omega\left(\frac{1/\gamma}{\poly\left(\log\left(1/\gamma\right)\right)}\right)$,
essentially matching the best known upper bound.
\end{abstract}

\newpage

\section{Introduction}
Let $\cx$ be some set and let $\cd$ be a distribution on $\cx\times
\{\pm 1\}$. The basic learning task is, based on an i.i.d. sample, to
find a function $f:\cx\to\{\pm 1\}$ whose error,
\mbox{$\Err_{\cd,0-1}(f):=\Pr_{(X,Y)\sim \cd}\left(f(X)\ne Y\right)$}, is as
small as possible. A {\em learning problem} is defined by specifying a
class $H$ of competitors (e.g. $H$ is a class of functions from $\cx$
to $\{\pm 1\}$). Given such a class, the corresponding learning
problem is to find $f:\cx\to\{\pm 1\}$ whose error is small relatively
to the error of the best competitor in $H$.  Ignoring
computational aspects, the celebrated PAC/VC theory essentially tells
us that the best algorithm for every learning problem is an Empirical
Risk Minimizer (=ERM) -- namely, one that returns the competitor in
$H$ of least {\em empirical error}. Unfortunately, for many learning
problems, implementing the ERM paradigm is $NP$-hard and even
$NP$-hard to approximate.

We consider here a very popular family of algorithms to cope with this hardness, which we collectively call {\em ``the generalized linear family"}. It proceeds as follows: fix some set $W\subset \reals^{\cx}$ and return a function of the form $f(x)=\sign(g(x)-b)$ where the pair $(g,b)\in W\times \reals$ empirically minimizes some convex loss. In order that such a method be useful, the set $W$ should be ``small" (to prevent overfitting) and ``nicely behaved'' (to make the optimization problem computationally feasible). The two main choices for such a set $W$ are
\begin{itemize}
\item A (usually convex) subset of a finite dimensional space of functions (e.g. if $\cx\subset \reals^d$ then $W$ can be the space of all polynomials of degree $\le 17$ and coefficients bounded by $d^3$). We refer to such algorithms as {\em finite dimensional learners}.
\item A ball in a reproducing kernel Hilbet space. We refer to such algorithms as as {\em kernel based learners}.
\end{itemize}

The generalized linear family has been applied extensively to tackle learning problems (e.g. \cite{LinialMaNi89, KushilevitzMa91,KlivansServedio01, KalaiKlMaSe05, BlaisOdWi08, ShalevShSr11} -- see section \ref{sec:related_work}). Their statistical charactersitics have been thoroughly studied as well \citep{Vapnik98, AnthonyBa99, ScholkopfBuSm98, CristianiniSh00, Steinwart08}. Moreover, the significance of this approach is by no means only theoretical -- algorithms from this family are widely used by practitioners.

In spite of all that, very few lower bounds are known on the performance of this family of algorithms (i.e., theorems of the form ``For every kernel-based/finite-dimensional algorithm for the learning problem $X$, there exists a distribution under which the algorithm performs poorly"). Such a lower bound must quantify over all possible choices of ``small and nicely behaved" sets $W$. In order to address this difficulty we employ a variety of mathematical methods some of which are new in this domain. In particular, we make intensive use of harmonic analysis on the sphere, reproducing kernel Hilbert spaces, orthogonal polynomials, John's Lemma as well as a new symmetrization technique.

We also prove a new result, which is of independent interest: a fundamental fact that stands behind the theoretical analysis of kernel based learners is that for every subset $\cx$ of a unit ball in a Hilbert space $H$, it is possible to learn affine functionals of norm $\le C$ over $\cx$ w.r.t. the hinge loss using $\frac{C^2}{\epsilon^2}$ examples. We show a (weak) inverse of this fact. Namely, we show that for every $\cx$, if affine functionals can be learnt using $m$ examples, then there exists an equivalent inner product on $H$ under which $\cx$ is contained in a unit ball, and the affine functional retuned by any learning algorithm must have norm $\le O\left(m^3\right)$.

Our lower bounds are established for the basic problem of {\em
  learning large margin halfspaces} (to be defined precisely in
Section \ref{sec:LLMH}). The best known efficient (in
$\frac{1}{\gamma}$) algorithm for this problem \citep{BirnbaumSh12} is
a kernel based learner that achieves an approximation ratio of
$\frac{1/\gamma}{\sqrt{\log(1/\gamma)}}$. (We note, however, that this
approximation ratio was first obtained by \citep{LongSe11} using a
``boosting based" algorithm that does not belong to the generalized
linear family). The best known exact algorithm (that is,
$\alpha(\gamma)=1$), is also a kernel based learner and runs in time
$\exp\left(\Theta\left(\frac{1}{\gamma}\log\left(\frac{1}{\gamma}\right)\right)\right)$
\citep{ShalevShSr11}.

Our main results show that efficient kernel based learners cannot achieve better approximation ratio than $\Omega\left(\frac{1/\gamma}{\poly\left(\log\left(1/\gamma\right)\right)}\right)$, essentially matching the best known upper bound. Also, we show that efficient finite dimensional learners cannot achieve better approximation ratio than $\Omega\left(\frac{1/\sqrt{\gamma}}{\poly\left(\log\left(1/\gamma\right)\right)}\right)$. In addition we show that the running time of kernel based learners with approximation ratio of $\left(\frac{1}{\gamma}\right)^{1-\epsilon}$ as well as of finite dimensional learners with approximation ratio of $\left(\frac{1}{\gamma}\right)^{\frac{1}{2}-\epsilon}$ must be \emph{exponential} in $1/\gamma$.

Next, we formulate the problem of learning large margin halfspaces and survey some relevant background to motivate our definitions of kernel-based and finite dimensional learners given in Section \ref{sec:results}.
\subsection{Learning large margin halfspaces}\label{sec:LLMH}
We view $\reals^d$ as a subspace of the Hilbert space $H=\ell^2$
corresponding to the first $d$ coordinates. Since the notion of margin
is defined relative to a suitable scaling of the examples, we consider
throughout only distributions that are supported in the unit ball,
$B$, of $H$. Also, all the distributions we consider are supported in
$\reals^d$ for some $d<\infty$. We denote by $S^{d-1}$ the unit sphere
of $\reals^d$. 

It will be convenient to use {\em loss functions}. A loss function is any function $l:\reals\to [0,\infty)$.
Given a loss function $l$ and $f:B\to\reals$, we denote 
$\Err_{\D,l}(f) = \E_{(x,y) \sim \D}(l(yf(x)))$.
Two loss functions of particular relevance are the $0-1$ loss function, $l_{0-1}(x) = \begin{cases} 1 & x\le 0 \\ 0 & x>0\end{cases}$, and the $\gamma$-margin loss function, $l_\gamma(x) =
\begin{cases} 1 & x\le \gamma \\ 0 & x>\gamma\end{cases}$. We use
shorthands such as $\Err_{\D,0-1}$ instead of $\Err_{\D,l_{0-1}}$.

A halfspace, parameterized by $w \in B$ and $b \in \reals$, is the classifier $f(x) =
\sign(\Lambda_{w,b}(x))$, where $\Lambda_{w,b}(x): = \inner{w,x} + b$. Given a distribution $\D$ over $B \times
\{\pm 1\}$, the error rate of $\Lambda_{w,b}$ is
\[
\Err_{\D,0-1}(\Lambda_{w,b}) = \Pr_{(x,y)\sim \mathcal
  D}\left(\sign(\Lambda_{w,b}(x)) \ne y\right) = \Pr_{(x,y)\sim \mathcal
  D}\left( y \Lambda_{w,b}(x) \le 0\right)~.
\]
The $\gamma$-margin error rate of $\Lambda_{w,b}$ is
\[
\Err_{\D,\gamma}(\Lambda_{w,b}) = \Pr_{(x,y)\sim \mathcal D}\left( y \Lambda_{w,b}(x) \le \gamma \right) ~.
\]
Note that if $\|w\|=1$ then $|\Lambda_{w,b}(x)|$ is the distance of $x$ from the separating hyperplane. Therefore, the $\gamma$-margin error rate is the probability of $x$ to either be in the wrong side of the hyperplane or to be at a distance of at most $\gamma$ from the hyperplane.  The least $\gamma$-margin error rate of a halfspace classifier is denoted $\Err_\gamma(\D) = \min_{w \in B,b \in \reals} \Err_{\D,\gamma}(\Lambda_{w,b})$.

A learning algorithm receives $\gamma,\epsilon$ and access to
i.i.d. samples from $\D$.  The algorithm should return a classifier (which {\em need not} be an affine function).
We say that the algorithm has approximation ratio
$\alpha(\gamma)$ if for every $\gamma,\epsilon$ and for every distribution, it outputs
(w.h.p. over the i.i.d.  $\D$-samples) a classifier
with error rate $\le \alpha(\gamma)
\Err_\gamma(\D) + \epsilon$. An {\em efficient} algorithm 
uses $\poly(1/\gamma,1/\epsilon)$ samples, runs in time polynomial in the size of the sample\footnote{The size of a vector $x \in H$ is taken to be  the largest index $j$ for which $x_j \neq 0$.} and 
outputs a classifier $f$ such that $f(x)$ can be evaluated in time polynomial in the sample size.

\subsection{Kernel-SVM and kernel-based learners} \label{sec:kernelDefs}
The SVM paradigm, introduced by Vapnik is inspired by the idea of separation with margin.
For the reader's convenience we first describe the basic (kernel-free) variant of SVM. It is well known (e.g. \cite{AnthonyBa99}) that the affine function that minimizes the \emph{empirical} $\gamma$-margin error rate over an i.i.d. sample of size $\poly(1/\gamma,1/\epsilon)$ has error rate  $\le\Err_\gamma(\D) + \epsilon$. However, this minimization problem is $NP$-hard and even $NP$-hard to approximate \citep{GuruswamiRa06, FeldmanGoKhPo06}.

SVM deals with this hardness by replacing the margin
loss with a {\em convex surrogate loss}, in particular, the {\em
  hinge loss}\footnote{As usual, $z_+:=\max(z,0)$.} $l_{\hinge}(x)=(1-x)_+$. Note that for $x \in [-2,2]$,
\[
l_{0-1}(x) \le l_{\hinge}(x/\gamma) \le (1+2/\gamma) l_\gamma(x) ~,
\]
from which it easily follows that by solving
\[
\min_{w,b} ~ \Err_{\mathcal D,\hinge}\left(\tfrac{1}{\gamma}\,\Lambda_{w,b}\right) 
~~\text{s.t.} ~~~ w\in H,\;b\in\reals, ~~\|w\|_{H}\le 1
\]
we obtain an approximation ratio of $\alpha(\gamma) = 1+2/\gamma$. 
It is more convenient to consider the problem 
\begin{equation}\label{eq:svm_0_hinge}
\min_{w,b} ~ \Err_{\mathcal D,\hinge}\left(\Lambda_{w,b}\right) 
~~\text{s.t.} ~~~ w\in H,\;b\in\reals, ~~\|w\|_{H}\le C ~,
\end{equation}
which is equivalent for $C = \frac{1}{\gamma}$. The basic (kernel-free) variant of SVM essentially solves Problem \eqref{eq:svm_0_hinge}, which can be approximated, up to an additive error of $\epsilon$, by an efficient algorithm running on a sample of size $\poly(\frac{1}{\gamma},\frac{1}{\epsilon})$.

Kernel-free SVM minimizes the hinge loss over the space of affine
functionals of bounded norm. The family of Kernel-SVM algorithms is
obtained by replacing the space of affine functionals with other,
possibly much larger, spaces (e.g., a polynomial kernel of degree $t$
extends the repertoire of possible output functions from affine
functionals to all polynomials of degree at most $t$). This is accomplished by embedding $B$ into the unit ball of another Hilbert space on which we apply basic-SVM. 
Concretely, let $\psi:B\to B_1$, where $B_1$ is the unit ball of a
Hilbert space $H_1$. The embedding $\psi$ need not be computed
directly. Rather, it is enough that we can efficiently compute the corresponding
{\em kernel}, $k(x,y):=\langle \psi(x),\psi(y)\rangle_{H_1}$ (this property, sometimes crucial, is called the {\em kernel trick}). It
remains to solve the following problem
\begin{equation}\label{eq:svm_1_hinge}
\min_{w,b}~  \Err_{\mathcal D,\hinge}\left(\Lambda_{w,b}\circ\psi\right) 
~~~\text{s.t.}~~~ w\in H_1,\;b\in\reals, ~\|w\|_{H_1}\le C ~.
\end{equation}
This problem can be approximated, up to an additive error of
$\epsilon$, using $\poly(C/\epsilon)$ samples and time. We prove lower bounds to all
approximate solutions of program (\ref{eq:svm_1_hinge}). In fact, our results work with arbitrary (not just hinge loss) convex surrogate losses and arbitrary (not just efficiently computable) kernels.

Although we formulate our results for Problem (\ref{eq:svm_1_hinge}), they apply as well to the following commonly used formulation of the kernel SVM problem, where the constraint $\|w\|_{H_1} \le C$ is replaced by a regularization term. Namely
\begin{equation} \label{eq:svm_2_hinge}
\min_{w \in H_1,b \in \reals}~  \frac{1}{C^2} \|w\|_{H_1}^2 + \Err_{\mathcal D,\hinge}\left(\Lambda_{w,b}\circ\psi\right) 
\end{equation}

The optimum of program (\ref{eq:svm_2_hinge}) is $\le 1$ as shown by
the zero solution $w=0,b=0$. Thus, if $w,b$ is an approximate optimal solution, then $\frac{\|w\|_{H_1}^2}{C^2}\le 2 \Rightarrow \|w\|_{H_1}\le 2C$. This observation makes it easy to modify our results on program (\ref{eq:svm_1_hinge}) to apply to program (\ref{eq:svm_2_hinge}).

\subsection{Finite dimensional learners}
The SVM algorithms embed the data in a (possibly infinite dimensional) Hilbert space, and minimize the hinge loss over all affine functionals of bounded norm. The kernel trick sometimes allows us to work in infinite dimensional Hilbert spaces. Even without it, we can still embed the data in $\reals^m$ for some $m$, and minimize a convex loss over a collection of affine functionals.
For example, some algorithms do not constraint the affine functional, while in the Lasso method~\citep{Tibshirani96b} the affine functional (represented as a vector in $\reals^m$) must have small $L^1$-norm. 

Without the kernel trick, such algorithms work directly in $\reals^m$. Thus, every algorithm must have time complexity $\Omega(m)$, and therefore $m$ is a lower bound on the complexity of the algorithm. In this work we will lower bound the performance of any algorithm with $m\le \poly\left(1/\gamma\right)$.  
Concretely, we prove lower bounds for any approximate solution to a problem of the form
\begin{equation}\label{eq:svm_finite_dim}
\min_{w,b}~\Err_{\mathcal D,l}\left(\Lambda_{w,b}\circ\psi\right) ~~~\textrm{s.t.}~~~
w \in W \subset \reals^m, ~ b \in \reals ~~,
\end{equation}
where $l$ is some surrogate loss function (see formal definition in
the next section) and \mbox{$\psi:B\to\reals^m$}. 

It is not hard to see that for any $m$-dimensional space $V$ of functions over the ball, there exists an embedding $\psi:B\to\reals^m$ such that
$$\{f+b:f\in V,b\in \reals\}=\{\Lambda_{w,b}\circ \psi:w\in \reals^m,b\in\reals\}$$
Hence, our lower bounds hold for any
method that optimizes a surrogate loss over a subset of a finite
dimensional space of functions, and return the threshold function corresponding to the optimum.

\subsection{Previous Results and Related Work}\label{sec:related_work}

The problem of learning halfspaces and in particular large margin halfspaces is as old as the field of machine learning, starting with the perceptron algorithm \citep{Rosenblatt58}. Since then it has been a fundamental challenge in machine learning and has inspired much of the existing theory as well as many popular algorithms.

The generalized linear method has its roots in the work of Gauss and
Legendre who used the least squares method for astronomical
computations. This method has played a key role in modern
statistics. Its first application in computational learning theory is
in \citep{LinialMaNi89} where it is shown that $AC^0$ functions are
learnable in quasi-polynomial time w.r.t. the uniform
distribution. Subsequently, many authors have used the method to
tackle various learning problems. For example,
\cite{KlivansServedio01} derived the fastest algorithm for learning
DNF and \cite{KushilevitzMa91} used it to develop an algorithm for
decision trees. The main uses of the linear method in the problem of
learning halfspaces appear in the next paragraph. Needless to say we
are unable here to offer a comprehensive survey of its uses in
computational learning theory in general.

The best currently known approximation ratios in the problem of
learning large margin halfspaces are due to \citep{BirnbaumSh12} and
\citep{LongSe11} and achieve an approximation ratio of
$\frac{1}{\gamma\cdot\sqrt{\log(1/\gamma)}}$. The algorithm of
\citep{BirnbaumSh12} is a kernel based learner, while \citep{LongSe11}
used a ``boosting based" approach (that does not belong to the
generalized linear method).  The fastest exact algorithm is due to
\cite{ShalevShSr11} and runs it time
$\exp\left(\Theta\left(\frac{1}{\gamma}\log\left(\frac{1}{\epsilon\gamma}\right)\right)\right)$,
and is also a kernel based learner.  Better running times can be
achieved under distributional assumptions.  For data which is
separable with margin $\gamma$, i.e. $\Err_\gamma(\D)=0$, the
perceptron algorithm (as well as SVM with a linear kernel) can find a
classifier with error $\le\epsilon$ with time and sample complexity
$\le \poly(1/\gamma,1/\epsilon)$.  \cite{KalaiKlMaSe05} gave a finite
dimensional learner which is the fastest known algorithm for learning
halfspaces w.r.t. the uniform distribution over $S^{d-1}$ and the
$d$-dimensional boolean cube (running in time
$d^{O\left(1/\epsilon\right)^4}$). They also designed a finite
dimensional learner of halfspaces w.r.t. log-concave
distributions. \cite{BlaisOdWi08} extended these results from uniform
to product distributions. In this work, we focus on algorithms which
work for any distribution and whose runtime is polynomial in both
$1/\gamma$ and $1/\epsilon$.

The problem of proper\footnote{A proper learner must output a
  halfspace classifier. Here we consider improper learning where the
  learner can output any classifier.} learning of halfspaces in the
non-separable case was shown to be hard to approximate within any
constant approximation factor \citep{FeldmanGoKhPo06,GuruswamiRa06}.
It has been recently shown \citep{ShalevShSr11} that improper learning
under the margin assumption is also hard (under some cryptographic
assumptions).  Namely, no polynomial time algorithm can achieve an
approximation ratio of $\alpha(\gamma)=1$. In another recent result \cite{daniely2013average} have shown that under a certain complexity assumption, for every constant $\alpha$, no polynomial time algorithm can achieve an
approximation ratio of $\alpha$.

\cite{Ben-DavidLoSreShr} (see also \cite{LongSe11}) addressed the
performance of methods that minimize a convex loss over the class of
affine functional of bounded norm (in our terminology, they considered
the narrow class of finite dimensional learners that optimize over the
space of linear functionals). They showed that the best approximation
ratio of such methods is $\Theta(1/\gamma)$. Our results can be seen
as a substantial generalization of their results.

The learning theory literature contains consistency results for
learning with the so-called universal kernels and well-calibrated
surrogate loss functions. This includes the study of asymptotic
relations between surrogate convex loss functions and the 0-1 loss
function \citep{Zhang04a,BartlettJoMc06,Steinwart08}. It is shown that
the approximation ratio of SVM with a universal kernel tends to $1$ as
the sample size grows. Our result implies that this convergence is
very slow, e.g., an exponentially large (in $\frac 1 {\gamma}$) sample
is needed to make the error $< 2\Err_{\gamma}(\D)$.

Also related are \cite{ben2003limitations} and \cite{warmuth2005leaving}. These papers show the existence of learning problems with limitations on the ability to learn them using linear methods.

\section{Results}\label{sec:results}
We first define the two families of algorithms to which our lower
bounds apply. We start with the class of {\em surrogate loss}
functions. This class includes the most popular choices such as the
absolute loss $|1-x|$, the squared loss $(1-x)^2$, the logistic loss
$\log_2\left(1+e^{-x}\right)$, the hinge loss $(1-x)_+$ etc.

\begin{definition}[Surrogate loss function]
A function $l : \reals \to \reals$ is called a surrogate loss function
if $l$ is convex and is bounded below by the 0-1 loss. 
\end{definition}
The first family of algorithms contains kernel based algorithms, such
as kernel SVM. In the definitions below we set the accuracy
parameter $\epsilon$ to be $\sqrt{\gamma}$. Since our goal is to prove
lower bounds, this choice is without loss of generality, and is
intended for the sake of simplifying the theorems statements. 
\begin{definition}[Kernel based learner]
Let $l : \reals \to \reals$ be a surrogate loss function. 
A kernel based learning algorithm, $A$, receives as input $\gamma \in
(0,1)$. It then selects $C=C_A(\gamma)$ and an absolutely continuous
feature mapping, $\psi=\psi_A(\gamma)$, which maps the original space
$H$ into the unit ball of a new space $H_1$ (see Section \ref{sec:kernelDefs}). The algorithm returns a function
$$A(\gamma)\in \{\Lambda_{w,b}\circ\psi:w\in H_1,b\in\reals,\|w\|_{H_1}\le C\}$$
such that, with probability $\ge 1-\exp(-1/\gamma)$,
$$\Err_{\cd,l}(A(\gamma))\le \inf\{\Err_{\cd,l}(\Lambda_{w,b}\circ\psi):w\in H_1,b\in\reals,\|w\|_{H_1}\le C\}+\sqrt{\gamma} ~.$$
We denote by $m_A(\gamma)$ the maximal number of examples $A$ uses.
We say that $A$ is {\em efficient} if $m_A(\gamma) \le \poly(1/\gamma)$. 
\end{definition}

Note that the definition of kernel based learner allows for any predefined convex surrogate loss, not just the hinge loss. Namely,
we consider the program
\begin{equation}\label{eq:svm_1_general}
\min_{w,b}~  \Err_{\mathcal D,l}\left(\Lambda_{w,b}\circ\psi\right) 
~~~\text{s.t.}~~~ w\in H_1,\;b\in\reals, ~\|w\|_{H_1}\le C ~.
\end{equation}
We note that our results hold even if the
kernel corresponds to $\psi$ is hard to compute. 

The second family of learning algorithms involves an arbitrary feature mapping and domain constraint on the vector $w$, as in program (\ref{eq:svm_finite_dim}). 

\begin{definition}[Finite dimensional learner]
  Let $l : \reals \to \reals$ be some surrogate loss function.  A
  finite dimensional learning algorithm, $A$, receives as input
  $\gamma \in (0,1)$. It then selects a continuous embedding
  $\psi=\psi_A(\gamma):B \to\reals^m$ and a constraint set
  $W=W_A(\gamma)\subseteq \reals^m$. The algorithm returns, with
  probability $\ge 1-\exp(-1/\gamma)$, a function
$$A(\gamma)\in \{\Lambda_{w,b}\circ\psi:w\in W,b\in\reals\}$$
such that
$$\Err_{\cd,l}(A(\gamma))\le \inf\{\Err_{\cd,l}(\Lambda_{w,b}\circ\psi):w\in W,b\in\reals\}+\sqrt{\gamma} ~.,$$
We say that $A$ is {\em efficient} if $m=m_A(\gamma) \le \poly(1/\gamma)$. 
\end{definition}

\subsection{Main Results}
We begin with a lower bound on the performance of efficient kernel-based algorithms.
\begin{theorem}\label{thm:main_1_simple}
Let $l$ be an arbitrary surrogate loss and let $A$ be an efficient
kernel-based learner w.r.t. $l$. Then, for every $\gamma>0$, there exists a distribution $\mathcal D$ on $B$ such that, w.p. $\ge 1-10\exp(-1/\gamma)$, 
\[\frac{\Err_{\cd,0-1}(A(\gamma))}{\Err_\gamma(\mathcal D)} \ge \Omega\left(\frac{1}{\gamma \cdot \operatorname{poly}(\log(1/\gamma))}\right)
~.
\]
\end{theorem}
Next we show that kernel-based learners that achieve approximation ratio of $\left(\frac{1}{\gamma}\right)^{1-\epsilon}$ for some constant $\epsilon>0$ must suffer exponential complexity.
\begin{theorem}\label{thm:main_1_constant}
Let $l$ be an arbitrary surrogate loss, let $\epsilon>0$ and let $A$ be a
kernel-based learner w.r.t. $l$ such that for every $\gamma>0$ and every distribution $\mathcal D$ on $B$, w.p. $\ge 1/2$, 
\[\frac{\Err_{\cd,0-1}(A(\gamma))}{\Err_\gamma(\mathcal D)} \le 
\left(\frac{1}{\gamma}\right)^{1-\epsilon} ~.
\]
Then, for some $a=a(\epsilon)>0$, $m_A(\gamma)=\Omega\left(\exp\left(\left(1/\gamma\right)^a\right)\right)$.
\end{theorem}
These two theorems follow from the following result.
\begin{theorem}\label{thm:main_1}
Let $l$ be an arbitrary surrogate loss and let $A$ be a
kernel-based learner w.r.t. $l$ for which $m_A(\gamma) = \exp(o(\gamma^{-2/7}))$. Then, for every $\gamma>0$, there exists a distribution $\mathcal D$ on $B$ such that, w.p. $\ge 1-10\exp(-1/\gamma)$, 
\[\frac{\Err_{\cd,0-1}(A(\gamma))}{\Err_\gamma(\mathcal D)} \ge \Omega\left(\frac{1}{\gamma \cdot \operatorname{poly}(\log(m_A(\gamma)))}\right)
~.
\]
\end{theorem}
It is shown in \citep{BirnbaumSh12} that solving kernel SVM with some specific
kernel (i.e. some specific $\psi$) yields an approximation ratio of
$O\left( \frac{1}{\gamma \sqrt{\log(1/\gamma)}} \right)$. It follows
that our lower bound in Theorem \ref{thm:main_1} is essentially tight.
Also, this theorem can be viewed as a substantial generalization of
\citep{Ben-DavidLoSreShr,LongSe11}, who give an approximation ratio of
$\Omega\left(\frac{1}{\gamma}\right)$ with no embedding (i.e., $\psi$
is the identity map). Also relevant is \citep{ShalevShSr11}, which shows that
for a certain $\psi$, and
$m_A(\gamma)=\poly\left(\exp\left((1/\gamma)\cdot\log\left(1/(\gamma\right)\right)\right)$,
kernel SVM has approximation ratio of $1$. Theorem \ref{thm:main_1}
shows that for kernel-based learner to achieve a constant
approximation ratio, $m_A$ must be exponential in $1/\gamma$.

Next we give lower bounds on the performance of finite dimensional learners.
\begin{theorem}\label{thm:main_1_fin_dim}
  Let $l$ be a Lipschitz surrogate loss and let $A$ be a finite
  dimensional learner w.r.t. $l$. Assume that $m_A(\gamma) =
  \exp(o(\gamma^{-1/8}))$.  Then, for every $\gamma>0$, there exists a
  distribution $\mathcal D$ on $S^{d-1}\times \{\pm 1\}$ with
  $d=O(\log(m_A(\gamma)/\gamma))$ such that, w.p. $\ge
  1-\exp(-1/\gamma)$,
\[
\frac{\Err_{\cd,0-1}(A(\gamma))}{\Err_\gamma(\mathcal D)} \ge 
\Omega\left(\frac{1}{\sqrt{\gamma}\poly(\log(m_A(\gamma)/\gamma))}\right) ~.
\]
\end{theorem}

\begin{corollary}\label{cor:main_1_fin_dim}
Let $l$ be a Lipschitz surrogate loss and let $A$ be an efficient finite
  dimensional learner w.r.t. $l$. Then, for every $\gamma>0$, there exists a distribution $\mathcal D$ on $S^{d-1}\times \{\pm 1\}$ with \mbox{$d=O(\log(1/\gamma))$} such that, w.p. $\ge 1-\exp(-1/\gamma)$,
\[
\frac{\Err_{\cd,0-1}(A(\gamma))}{\Err_\gamma(\mathcal D)} \ge 
\Omega\left(\frac{1}{\sqrt{\gamma}\poly(\log(1/\gamma))}\right) ~.
\]
\end{corollary}
\begin{corollary}\label{thm:main_1_fin_constant}
Let $l$ be a Lipschitz surrogate loss, let $\epsilon>0$ and let $A$ be a finite
  dimensional learner w.r.t. $l$ such that for every $\gamma>0$ and every distribution $\mathcal D$ on $B^d$ with $d=\omega(\log(1/\gamma))$ it holds that w.p. $\ge 1/2$, 
\[\frac{\Err_{\cd,0-1}(A(\gamma))}{\Err_\gamma(\mathcal D)} \le \left(\frac{1}{\gamma}\right)^{\frac{1}{2}-\epsilon}
\]
Then, for some $a=a(\epsilon)>0$, $m_A(\gamma)=\Omega\left(\exp\left(\left(1/\gamma\right)^a\right)\right)$.
\end{corollary}
\subsection{Review of the proofs' main ideas}

To give the reader some idea of our arguments, we sketch some of the
main ingredients of the proof of Theorem \ref{thm:main_1}. At the end of this section we sketch the idea of the proof of Theorem
\ref{thm:main_1_fin_dim}. We note, however, that
the actual proofs are organized somewhat differently.

We will construct a distribution $\D$ over $S^{d-1} \times \{\pm 1\}$
(recall that $\reals^d$ is viewed as standardly embedded in $H=\ell^2$). Thus, we can assume that the program is
formulated in terms of the unit sphere, $S^{\infty} \subset \ell^2$,
and not the unit ball.

Fix an embedding $\psi$ and $C>0$. Denote by $k:S^{\infty}\times
S^{\infty}\to\reals$ the corresponding kernel $k(x,y)=\langle
\psi(x),\psi(y)\rangle_{H_1}$ and consider the following set of
functions over $S^{\infty}$
$$H_k=\{\Lambda_{v,0}\circ \psi: v\in H_1 \} ~.$$
$H_k$ is a Hilbert space with norm
$||f||_{H_k}=\inf\{||v||_{H_1}:\Lambda_{v,0}\circ \psi=f\}$. The
subscript $k$ indicates that $H_k$ is uniquely determined (as a
Hilbert space) given the kernel $k$. With this interpretation, program
(\ref{eq:svm_1_general}) is equivalent to the program
\begin{equation}\label{eq:svm_outline}
\min_{f \in H_k,b \in \reals}  \Err_{\mathcal D,l}\left(f+b\right) 
~~\text{s.t.}~~ ||f||_{H_k}\le C ~.
\end{equation}
For simplicity we focus on $l$ being the hinge-loss (the
generalization to other surrogate loss functions is rather technical).

The proof may be split into four steps:
\begin{enumerate}
\item Our first step is to show that we can restrict to the case $C=\poly\left(\frac{1}{\gamma}\right)$. We show that for every subset $\cx$ of a Hilbert space $H$, if affine functionals on $\cx$ can be learnt using $m$ examples w.r.t. the hinge loss, then there exists an equivalent inner product on $H$ under which $\cx$ is contained in a unit ball, and the affine functional returned by any learning algorithm must have norm $\le O\left(m^3\right)$. Since we consider algorithms with polynomial sample complexity, this allows us to argue as if $C=\poly\left(\frac{1}{\gamma}\right)$.
\item We consider the one-dimensional problem of
  improperly learning halfspaces (i.e. thresholds on the line) by
  optimizing the hinge loss over the space of univariate polynomials of
  degree bounded by $\log(C)$. We construct a distribution $\cd$ over
  $[-1,1]\times \{\pm 1\}$ that is a convex combination of two distributions. One that is separable by a $\gamma$-margin halfspace and the other representing a tiny amount of noise.
  We show that each solution of the problem of minimizing the hinge-loss w.r.t. $\cd$ over the
  space of such polynomials has the property that $f(\gamma)
  \approx f(-\gamma) $. 
\item We pull back the distribution $\cd$ w.r.t. a direction
  $e \in S^{d-1}$ to a distribution over $S^{d-1} \times \{\pm 1\}$.
Let $f$ be an approximate solution of program (\ref{eq:svm_outline}). We show that $f$ takes almost the same value on instances for
  which $\inner{x,e}=\gamma$ and $\inner{x,e} = -\gamma$. This step can be further broken into three substeps --
\begin{enumerate}
\item First, we assume that the kernel is symmetric and $f(x)$ depends only on $\inner{x,e}$. This substep uses a characterization of Hilbert spaces corresponding to symmetric kernels, from which it follows that $f$ has the form
\[
f(x)=\sum_{n=1}^\infty \alpha_nP_{d,n}(\inner{x,e})~.
\]  
Here $P_{d,n}$ are the $d$-dimensional Legendre polynomials and
$\sum_{n=0}^\infty \alpha_n^2< C^2$. This allows us to rely on the results
for the one-dimensional case from step (1).
\item By symmetrizing $f$, we relax the assumption that $f$ depends only on $\inner{x,e}$.
\item By averaging the kernel over the group of linear isometries on $\reals^d$, we relax the assumption that the kernel is symmetric. 
\end{enumerate}

\item Finally, we show that for the distribution from the previous step, if $f$ is an approximate solution to program (\ref{eq:svm_outline}) then $f$ predicts the same value, $1$, on instances for
  which $\inner{x,e}=\gamma$ and $\inner{x,e} = -\gamma$. This establishes our claim, as the constructed distribution assigns the value $-1$ to instances for which $\inner{x,e}=-\gamma$. 
\end{enumerate}

We now expand on this brief description of the main steps.

\subsubsection*{Polynomial sample implies small $C$} 
Let $\cx$ be a subset of the unit ball of some Hilbert space $H$ and let $C>0$. Assume that affine functionals over $\cx$ with norm $\le C$ can be learnt using $m$ examples with error $\epsilon$ and confidence $\delta$. That is, assume that there is an algorithm such that
\begin{itemize}
\item Its input is a sample of $m$ points in $\cx\times\{\pm 1\}$ and its output is an affine functional $\Lambda_{w,b}$ with $\|w\|\le C$.
\item For every  distribution $\cd$ on $\cx\times\{\pm 1\}$, it returns, with probability $1-\delta$, $w,b$ with $\Err_{\cd,\hinge}(\Lambda_{w,b})\le \inf_{\|w'\|\le C,b'\in\mathbb R}\Err_{\cd,\hinge}(\Lambda_{w',b'})+\epsilon$.
\end{itemize}
We will show that there is an equivalent inner product $\inner{\cdot,\cdot}'$ on $H$ under which $\cx$ is contained in a unit ball (not necessarily around $0$) and the affine functional returned by any learning algorithm as above, must have norm $\le O\left(m^3\right)$ w.r.t. the new norm.

The construction of the norm is done as follows. We first find an affine subspace $M\subset H$ of dimension $d\le m$ that is very close to $\cx$ in the sense that the distance of every point in $\cx$ from $M$ is $\le\frac{m}{C}$. To find such an $M$, we assume toward a contradiction that there is no such $M$, and use this to show that there is a subset $A\subset \cx$ such that every function $f:A\to [-1,1]$ can be realized by some affine functional with norm $\le C$. This contradicts the assumption that affine functionals with norm $\le C$ can be learnt using $m$ examples.

Having the subspace $M$ at hand, we construct, using John's lemma (e.g. \cite{Matousek02}), an inner product $\inner{\cdot,\cdot}''$ on $M$ and a distribution $\mu_N$ on $\cx$ with the property that the projection of $\cx$ on $M$ is contained in a ball of radius $\frac{1}{2}$ w.r.t. $\inner{\cdot,\cdot}''$, and the hinge error of every affine functional w.r.t. $\mu_N$ is lower bounded by the norm of the affine functional, divided by $m^2$.
We show that that this entails that any affine functional returned by the algorithm must have a norm $\le O\left(m^3\right)$ w.r.t. the inner product $\inner{\cdot,\cdot}''$. 

Finally, we construct an inner product $\inner{\cdot,\cdot}'$ on $H$ by putting the norm $\inner{\cdot,\cdot}''$ on $M$ and multiplying the original inner product by $\frac{C^2}{2m^2}$ on $M^{\perp}$.

\subsubsection*{The one dimensional distribution} 
We define a distribution $\D$ on $[-1,1]$ as follows. Start
with the distribution $\D_1$ that takes the values
$\pm(\gamma,1)$, where $\D_1(\gamma,1) = 0.7$ and $\D_1(-\gamma,-1) =
0.3$. Clearly, for this distribution, the threshold $0$ has zero error rate. To construct $\D$, we perturb $\D_1$ with ``noise'' as follows. 
Let $\D = (1-\lambda) \D_1 + \lambda \D_2$, where $\D_2$ is
defined as follows.  The probability of the labels is uniform and
independent of the instance and the marginal probability over the
instances is defined by the density function
\[
\rho(x) = \begin{cases}
0 & \textrm{if}~ |x| > 1/8 \\
\frac{8}{\pi\sqrt{1-\left( 8x\right)^2}}& \textrm{if}~ |x| \le 1/8
\end{cases} ~~.
\]
This choice of $\rho$ simplifies our calculations due to its relation
to Chebyshev polynomials. However, other choices of $\rho$ which are
supported on a small interval around zero can also work.

Note that the error rate of the threshold $0$ on $\D$ is
$\lambda/2$. We next show that each polynomial $f$ of degree
$K=\log(C)$ that satisfies $\Err_{\D,\hinge}(f) \le 1$ must have
$f(\gamma) \approx f(-\gamma)$. 
Indeed, if
$$1 \ge \Err_{\mathcal D,\hinge}(f)=(1-\lambda)\Err_{\mathcal
  D_1,\hinge}(f)+\lambda\Err_{\mathcal D_2,\hinge}(f)$$
then $\Err_{\mathcal D_2,\hinge}(f)\le \frac{1}{\lambda}$. But, 
\begin{align*}
\Err_{\mathcal D_2,\hinge}(f) &= \frac{1}{2} \int_{-1}^{1} l_{\hinge}(f(x))
\rho(x) dx + \frac{1}{2} \int_{-1}^{1} l_{\hinge}( -f(x))
\rho(x) dx \\
&\ge \frac{1}{2}\int_{-1}^1 l_{\hinge}(-|f(x)|)
\rho(x) dx 
\end{align*}
and using the convexity of $l_{\hinge}$ we obtain from Jensen's
inequality that 
\begin{align*}
&\ge \frac{1}{2} l_{\hinge}\left(\int_{-1}^1 -|f(x)|
\rho(x) dx\right) \\
&= \frac{1}{2} \left(1 + \int_{-1}^1 |f(x)|
\rho(x) dx\right) \\
&\ge \frac{1}{2}\int_{-1}^1 |f(x)|
\rho(x) dx =: \frac{1}{2}\|f\|_{1,d\rho} ~.
\end{align*}
This shows that $\|f\|_{1,d\rho} \le \frac{2}{\lambda}$. We
next write $f = \sum_{i=1}^K \alpha_i \tilde{T}_i$, where
$\{\tilde{T}_i\}$ are the orthonormal polynomials corresponding to the
measure $d\rho$. Since $\tilde{T}_i$ are related to Chebyshev
polynomials we can uniformly bound their $\ell_\infty$ norm, hence
obtain that 
\[
\sqrt{\sum_i
  \alpha_i^2} = \|f\|_{2,d\rho} \le 
O(\sqrt{K})\,\|f\|_{1,d\rho} \le
O\left(\frac{\sqrt{K}}{\lambda}\right) ~.
\]
Based on the above, and using a bound on the derivatives of Chebyshev
polynomials, we can bound the derivative of the polynomial $f$
\[
|f'(x)| \le \sum_i |\alpha_i| |\tilde{T}'_i(x)| \le 
O\left(\frac{K^3}{\lambda}\right) ~.
\]
Hence, by choosing $\lambda = \omega(\gamma K^3) = \omega(\gamma
\log^3(C))$ we obtain
\[
|f(\gamma)-f(-\gamma)| \le 2 \,\gamma \max_x |f'(x)| =
O\left(\frac{\gamma\,K^3}{\lambda}\right)  = o(1) ~,
\]
as required.

\subsubsection*{Pulling back to the $d-1$ dimensional sphere}

Given the distribution $\D$ over $[-1,1]\times \{\pm 1\}$ described
before, and some $e\in S^{d-1}$, we now define a distribution
$\D_e$ on $S^{d-1}\times \{\pm 1\}$. To sample from $\D_e$, we first sample $(\alpha, \beta)$ from $\mathcal D$
and (uniformly and independently) a vector $z$ from the
$1$-codimensional sphere of $S^{d-1}$ that is orthogonal to $e$. The
constructed point is $(\alpha e+\sqrt{1-\alpha^2}z,\beta)$. 

For any $f \in H_k$ and $a \in [-1,1]$ define $\bar{f}(a)$ to be the
expectation of $f$ over the $1$-codimensional sphere $\{x \in S^{d-1}
: \inner{x,e}=a\}$. We will show that for any $f \in H_k$, such that $\|f\|_{H_k} \le C$ and
$\Err_{\D_e,\hinge}(f) \le 1$, we have that $|\bar{f}(\gamma)-
\bar{f}(-\gamma)| = o(1)$. 

To do so, let us first assume that $f$ is symmetric with respect to
$e$, and hence can be written as
\[
f(x) = \sum_{n=0}^\infty \alpha_n P_{d,n}(\inner{x,e}) ~, 
\]
where $\alpha_n \in \reals$ and $P_{d,n}$ is the $d$-dimensional
Legendre polynomial of degree $n$. 
Furthermore, by a characterization of Hilbert spaces corresponding to symmetric
kernels, it follows that $\sum \alpha_n^2 \le C^2$. 

Since $f$ is symmetric w.r.t. $e$ we have, 
\[
\bar{f}(a) = \sum_{n=0}^\infty \alpha_n P_{d,n}(a) ~. 
\]
For $|a| \le 1/8$, we have that $|P_{d,n}(a)|$ tends to zero
exponentially fast with both $d$ and $n$. Hence, if $d$ is large
enough then
\[
\bar{f}(a) \approx  \sum_{n=0}^{\log(C)} \alpha_n P_{d,n}(a) =: \tilde{f}(a)~. 
\]
Note that $\tilde{f}$ is a polynomial of degree bounded by
$\log(C)$. In addition, by construction, $\Err_{\D_e,\hinge}(f) = \Err_{\D,\hinge}(\bar{f})
\approx \Err_{\D,\hinge}(\tilde{f})$. Hence, if $1 \ge
\Err_{\D_e,\hinge}(f)$ then using the previous subsection we conclude
that $|\bar{f}(\gamma)-\bar{f}(-\gamma)| = o(1)$.

\subsubsection*{Symmetrization of $f$}

In the above, we assumed that both the kernel function is symmetric
and that $f$ is symmetric w.r.t. $e$. Our next step is to relax the
latter assumption, while still assuming that the kernel function is
symmetric. 

Let $\mathbb{O}(e)$ be the group of linear isometries that fix $e$, namely, $\mathbb{O}(e) = \{A \in \mathbb{O}(d) : Ae=e\}$. By assuming that $k$ is a symmetric
kernel, we have that for all $A \in \mathbb{O}(e)$, the function \mbox{$g(x)
= f(Ax)$} is also in $H_k$. Furthermore, $\|g\|_{H_k} = \|f\|_{H_k}$
and by the construction of $\D_e$ we also have $\Err_{\D_e,\hinge}(g)
= \Err_{\D_e,\hinge}(f)$. Let $\cp_ef(x)=\int_{\mathbb O(e)}f(Ax)dA$
be the symmetrization of $f$ w.r.t. $e$. On one hand, $\cp_ef \in
H_k$, $\|\cp_ef\|_{H_k} \le \|f\|_{H_k}$, and
$\Err_{\D_e,\hinge}(\cp_ef) \le \Err_{\D_e,\hinge}(f)$. On the other
hand, $\bar{f} = \overline{\cp_ef}$. Since for $\cp_ef$ we have already
shown that
$|\overline{\cp_ef}(\gamma)-\overline{\cp_ef}(-\gamma)|=o(1)$, it
follows that $|\bar{f}(\gamma)-\bar{f}(-\gamma)| = o(1)$ as well.

\subsubsection*{Symmetrization of the kernel}

Our final step is to remove the assumption that the kernel is
symmetric. To do so, we first symmetrize the kernel as follows. Recall
that $\mathbb O(d)$ is the group of linear isometries of
$\reals^d$. Define the following symmetric kernel:
\[
k_s(x,y) = \int_{\mathbb{O}(d)} k(Ax,Ay) dA~.
\]
We show that the corresponding Hilbert space consists of functions of
the form
\[
f(x) = \int_{\mathbb{O}(d)} f_A(Ax) dA ~,
\]
where for every $A$ $f_A \in H_k$. Moreover,
\begin{equation} \label{eq:norm_bound_ks}
\|f\|^2_{H_{k_s}} ~\le~  \int_{\mathbb{O}(d)} \|f_A\|^2_{H_k} dA ~.
\end{equation}
Let $\alpha$ be the maximal number such that
\[
\forall e \in S^{d-1} \exists f_e \in H_k ~\textrm{s.t.}~
\|f_e\|_{H_k} \le C,~\Err_{D_e,\hinge}(f_e) \le 1,~
|\bar{f_e}(\gamma)-\bar{f_e}(-\gamma)| > \alpha ~.
\]
Since $H_k$ is closed to negation, it follows that $\alpha$ satisfies 
\[
\forall e \in S^{d-1} \exists f_e \in H_k ~\textrm{s.t.}~
\|f_e\|_{H_k} \le C,~\Err_{D_e,\hinge}(f_e) \le 1,~
\bar{f_e}(\gamma)-\bar{f_e}(-\gamma) > \alpha ~.
\]
Fix some $v \in S^{d-1}$ and define $f \in H_{k_s}$ to be
\[
f(x) = \int_{\mathbb O(d)} f_{Av}(Ax) dA ~.
\]
By Equation (\ref{eq:norm_bound_ks}) we have that $\|f\|_{H_{k_s}} \le
C$. It is also possible to show that for all $A$
$\Err_{\D_v,\hinge}(f_{Av} \circ A) = \Err_{\D_{Av},\hinge}(f_{Av}
)\le 1$. Therefore, by the convexity of the loss,
$\Err_{\D_v,\hinge}(f) \le 1$. It follows, by the previous sections,
that $|\bar{f}(\gamma) - \bar{f}(-\gamma)| = o(1)$. But, 
we show that $\bar{f}(\gamma) - \bar{f}(-\gamma) > \alpha$. 
It therefore follows that $\alpha = o(1)$, as required.

\subsubsection*{Concluding the proof}

We have shown that for every kernel, there exists some direction $e$
such that for all $f \in H_k$ that satisfies $\|f\|_{H_k} \le C$ and
$\Err_{\D_e,\hinge}(f) \le 1$ we have that $|\bar{f}(\gamma) -
\bar{f}(-\gamma)|= o(1)$. 

Next, consider $f$ which is also an (approximated) optimal solution of
program (\ref{eq:svm_1_hinge}) with respect to $\D_e$. Since $\Err_{\D_e,\hinge}(0)
= 1$, we clearly have that $\Err_{\D_e,\hinge}(f) \le 1$, hence
$|\bar{f}(\gamma) - \bar{f}(-\gamma)|= o(1)$. Next we show that
$\bar{f}(-\gamma) > 1/2$, which will imply that $f$ predicts the label
$1$ for most instances on the $1$ co-dimensional sphere such that
$\inner{x,e}=-\gamma$. Hence, its 0-1 error is close to $0.3
(1-\lambda) \ge 0.2$ while $\Err_{\gamma}(\D_e) = \lambda/2$. By
choosing $\lambda = O(\gamma \log^{3.1}(C))$ we obtain that the
approximation ratio is $\Omega\left(\frac{1}{\gamma
    \log^{3.1}(C)}\right)$. 

It is therefore left to show that $\bar{f}(-\gamma) > 1/2$.  Let $a =
\bar{f}(\gamma) \approx \bar{f}(-\gamma)$.  On \mbox{$(1-\lambda)$} fraction
fraction of the distribution, the hinge-loss would be (on average and
roughly) \mbox{$0.3 [1 + a]_+ + 0.7 [1 - a]_+$}. This function is minimized
for $a=1$, which concludes our proof since $\lambda$ is $o(1)$.

\subsubsection*{The proof of Theorem \ref{thm:main_1_fin_dim}}
To prove Theorem \ref{thm:main_1_fin_dim}, we prove, using John's
Lemma \citep{Matousek02}, that for every embedding $\psi:S^{d-1}\to B_1$, we can construct a kernel $k:S^{d-1}\times S^{d-1}\to\reals$ and a 
probability measure $\mu_N$ over $S^{d-1}$ with the following properties: If $f$ is an approximate solution of program (\ref{eq:svm_finite_dim}), where $\gamma$ fraction of the distribution $\cd$ is perturbed by $\mu_N$, then $\|f\|_k\le O\left(\frac{m^{1.5}}{\gamma}\right)$. Using this, we adapt the proof as sketched above to prove Theorem \ref{thm:main_1_fin_dim}.

\section{Additional Results}
{\bf Low dimensional distributions.} It is of interest to examine Theorem \ref{thm:main_1} when $\mathcal D$ is supported on $B^d$ for $d$ small. We show that for $d= O(\log(1/\gamma))$, the approximation ratio is $\Omega\left(\frac{1}{\sqrt{\gamma}\cdot\poly(\log(1/\gamma))}\right)$. Most commonly used kernels (e.g., the polynomial, RBF, and Hyperbolic tangent kernels, as well as the kernel used in \citep{ShalevShSr11}) are symmetric. Namely, for all unit vectors $x,y\in B$, $k(x,y):=\langle \psi(x),\psi(y)\rangle_{H_1}$ depends only on $\langle x,y\rangle_{H}$. For symmetric kernels, we show that even with the restriction that $d= O(\log(1/\gamma))$, the approximation ratio is still $\Omega\left(\frac{1}{\gamma\cdot\poly(\log(1/\gamma))}\right)$. However, the result for symmetric kernels is only proved for (idealized) algorithms that return the exact solution of program (\ref{eq:svm_1_general}).

\begin{theorem}\label{thm:main_2}
Let $A$ be a kernel-based learner corresponding to a Lipschitz surrogate. Assume that $m_A(\gamma) = \exp(o(\gamma^{-1/8}))$. Then, for every $\gamma>0$, there exists a distribution $\mathcal D$ on $B^d$, for $d=O(\log(m_A(\gamma)/\gamma))$, such that, w.p. $\ge 1-\exp(-1/\gamma)$, 
\[\frac{\Err_{\cd,0-1}(A(\gamma))}{\Err_\gamma(\mathcal D)} \ge \Omega\left(\frac{1}{\sqrt{\gamma} \cdot \operatorname{poly}(\log(m_A(\gamma)))}\right)
~.
\]
\end{theorem}

\begin{theorem}\label{thm:main_3}
Assume that $m_A(\gamma)=\exp(o(\gamma^{-2/7}))$ and $\psi$ is continuous and symmetric. For every $\gamma>0$, there exists a distribution $\mathcal D$ on $B^d$, for $d=O(\log(m_A(\gamma)))$ and a solution to program (\ref{eq:svm_1_general}) whose 0-1-error is $\Omega\left(\frac{1}{\gamma\operatorname{poly}(\log(m_A(\gamma)))}\right)\cdot\Err_\gamma(\mathcal D)$.
\end{theorem}

{\bf The integrality gap.} In bounding the approximation ratio,
we considered a predefined loss $l$. We believe that similar
bounds hold as well for algorithms that can choose $l$ according to
$\gamma$. However, at the moment, we only know to lower bound the {\em
  integrality gap}, as defined below. 

If we let $l$ depend on $\gamma$, we should redefine the complexity of Program (\ref{eq:svm_1_general}) to be $C\cdot L$, where $L$ is the Lipschitz constant of $l$. (See the discussion following Program (\ref{eq:svm_1_general})). The {\em ($\gamma$-)integrality gap} of program (\ref{eq:svm_1_general}) and (\ref{eq:svm_finite_dim}) is defined as the worst case, over all possible choices of $\mathcal D$, of the ratio between the optimum of the program, running on the input $\gamma$, to $\Err_\gamma(\mathcal D)$. 
We note that $ \Err_{\mathcal D,0-1}(f)\le \Err_{\mathcal D,l}(f)$ for every convex surrogate $l$. Thus, the integrality gap always upper bounds the approximation ratio. Moreover, this fact establishes most (if not all) guarantees for algorithms that solve Program (\ref{eq:svm_1_general}) or Program (\ref{eq:svm_finite_dim}).

We denote by $\partial_+f$ the right derivative of the real function $f$.
Note that $\partial_+f$ always exists for $f$ convex. Also, $\forall x\in\reals,\;|\partial_+f(x)|\le L$ if $f$ is $L$-Lipschitz. We prove:
\begin{theorem}\label{thm:main_4}
Assume that $C=\exp\left(o(\gamma^{-2/7})\right)$ and $\psi$ is continuous. For every $\gamma>0$, there exists a distribution $\mathcal D$ on $B^d$, for $d=O(\log(C))$ such that the optimum of Program (\ref{eq:svm_1_general}) is $\Omega\left(\frac{1}{\gamma\operatorname{poly}(\log(C\cdot |\partial_+l(0)|))}\right)\cdot\Err_\gamma(\mathcal D)$.
\end{theorem}
Thus Program (\ref{eq:svm_1_general}) has itegrality gap $\Omega\left(\frac{1}{\gamma\operatorname{poly}(\log(C\cdot L))}\right)$. For Program (\ref{eq:svm_finite_dim}) we prove a similar lower bound:

\begin{theorem}\label{thm:main_4_fin_dim}
Let $m,d,\gamma$ such that $d=\omega(\log(m/\gamma))$ and $m= \exp\left(o(\gamma^{-2/7})\right)$. There exist a distribution $\mathcal D$ on $S^{d-1}\times \{\pm 1\}$ such that the optimum of Program (\ref{eq:svm_finite_dim}) is \mbox{$\Omega\left(\frac{1}{\gamma\poly(\log(m/\gamma))}\right)\cdot\Err_\gamma(\mathcal D)$}.
\end{theorem}

\section{Conclusion}
We prove impossibility results for the family
of generalized linear methods in the task of learning large margin halfspaces. Some of our lower bounds nearly
match the best known upper bounds and we conjecture that the rest of our bounds can be improved as well to match the best known upper bounds.
As we describe next, our work leaves much for future research.

First, regarding the task of learning large margin halfspaces, our analysis suggests that if
better approximation ratios are at all possible then they would require
methods other than optimizing a convex surrogate over a
regularized linear class of classifiers. 

Second, similar to the problem of learning large margin halfsapces,
for many learning problems the best known algorithms belong to the
generalized linear family. Understanding the limits of the generalized
linear method for these problems is therefore of particular interest
and might indicate where is the line discriminating between
feasibility and infeasibility for these problems.  We believe that our
techniques will prove useful in proving lower bounds on the
performance of generalized linear methods for these and other learning
problems.  E.g., our techniques yield lower bounds on the performance
of generalized linear algorithms that learn halfspaces over the
boolean cube $\{\pm 1\}^n$: it can be shown that these methods cannot
achieve approximation ratios better than $\tilde{\Omega}(\sqrt{n})$
even if the algorithm competes only with halfspaces defined by vectors
in $\{\pm 1\}^n$. These ideas will be elaborated on in a long version
of this manuscript. 

Third, while our results indicate the limitations of generalized
linear methods, it is an empirical fact that these methods perform
very well in practice. Therefore, it is of great interest to find
conditions on distributions that hold in practice, under which
these methods guaranteed to perform well. We note that learning
halfspaces under distributional assumptions, has already been
addressed to a certain degree. For example,
\citep{KalaiKlMaSe05,BlaisOdWi08} show positive results under several
assumptions on the marginal distribution (namely, they assume that the
distribution is either uniform, log-concave or a product
distribution). There is still much to do here, specifically in search
of better runtimes. Currently these results yield a runtime which is
exponential in $\poly(1/\epsilon)$, 
where $\epsilon$ is the excess error of the learnt hypothesis.

Fourth, as part of our proof, we have shown a (weak) inverse (lemma \ref{lem:C_is_small_main}) of the famous fact that affine functionals of norm $\le C$ can be learnt using $\poly(C)$ samples. We made no attempts to prove a quantitative optimal result in this vein, and we strongly believe that much sharper versions can be proved. This interesting direction is largely left as an open problem.

There are several limitations of our analysis that deserve further
work. In our work the surrogate loss is fixed in advance. We
believe that similar results hold even if the loss depends on
$\gamma$. This belief is supported by our results about the
integrality gap. As explained in Section \ref{sec:choosing_l}, this is
a subtle issue that related to questions about sample complexity.
Finally, in view of Theorems \ref{thm:main_4} and
\ref{thm:main_4_fin_dim}, we believe that, as in Theorem
\ref{thm:main_1}, the lower bound in Theorems \ref{thm:main_1_fin_dim}
and \ref{thm:main_2} can be improved to depend on $\frac{1}{\gamma}$
rather than on $\frac{1}{\sqrt{\gamma}}$.

\section{Proofs}
\subsection{Background and Notation}
Here we introduce some notations and terminology to be used throughout. The $L^p$ norm corresponding to a measure $\mu$ is denoted $||\cdot||_{p,\mu}$. Also, $\mathbb N=\{1,2,\ldots\}$ and $\mathbb N_0=\{0,1,2,\ldots\}$.
For a collection of function $\cf\subset\cy^{\cx}$ and $A\subset \cx$ we denote $\cf|_{A}=\{f|_A\mid f\in \cf\}$.
Let $H$ be a Hilbert space. We denote the projection on a closed convex subset of $H$ by $P_C$. We denote the norm of an affine functional $\Lambda$ on $H$ by $\|\Lambda\|_H=\sup_{\|x\|_H=1}|\Lambda(x)-\Lambda(0)|$.

\subsubsection{Reproducing Kernel Hilbert Spaces}
All the theorems we quote here are standard and can be found, e.g., in Chapter 2 of \citep{Saitoh88}.
Let $H$ be a Hilbert space of functions from a set $S$ to $\complex$.
Note that $H$ consists of functions and not of equivalence classes of functions. We say that $H$ is a {\em reproducing kernel Hilbert space} (RKHS for short) if, for every $x\in S$, the linear functional $f \to f(x)$ is bounded.

A function $k:S\times S\to \complex$ is a {\em reproducing kernel} (or just a {\em kernel}) if, for every $x_1,\ldots, x_n\in S$, the matrix $\{k(x_i,x_j)\}_{1\le i,j\le n}$ is positive semi-definite.

Kernels and RKHSs are essentially synonymous:
\begin{theorem}\label{thm:RKHS_basic}
\begin{enumerate}
\item For every kernel $k$ there exists a unique RKHS $H_k$ such that for every $y\in S$, $k(\cdot,y)\in H_k$ and $\forall f\in H,\;f(y)=\langle f(\cdot ),k(\cdot,y)\rangle_{H_k}$.
\item A Hilbert space $H\subseteq \complex^S$ is a RKHS if and only if there exists a kernel $k:S\times S\to \reals$ such that $H=H_k$.
\item For every kernel $k$, $\overline{\operatorname{span}\{k(\cdot,y)\}_{y\in S}}=H_k$. Moreover,
$$\langle \sum_{i=1}^n\alpha_ik(\cdot,x_i),\sum_{i=1}^n\beta_ik(\cdot,y_i)\rangle_{H_k}=\sum_{1\le i,j,\le n}\alpha_i\bar\beta_jk(y_j,x_i)$$
\item If the kernel $k:S\times S\to\reals$ takes only real values, then $H^\reals_k:=\{\operatorname{Re}(f):f\in H_k\}\subset H_k$. Moreover, $H^\reals_k$ is a real Hilbert space with the inner product induced from $H_k$.
\item For every kernel $k$, convergence in $H_k$ implies point-wise convergence. If $\sup_{x\in S}k(x,x)<\infty$ then this convergence is uniform.
\end{enumerate}
\end{theorem}\label{thm:RKHS_embedding}
There is also a tight connection between embeddings of $S$ into a Hilbert space and RKHSs.
\begin{theorem}
A function $k:S\times S\to\reals$ is a kernel iff there exists a mapping $\phi:S\to H$ to some real Hilbert space for which $k(x,y)=\langle \phi(y),\phi(x)\rangle_H$. Also,
$$H_k=\{f_v :v\in  H\}$$
Where $f_v(x)=\langle v,\phi (x)\rangle_{ H}$. The mapping $v\mapsto f_v$, restricted to $\overline{\operatorname{span}(\phi(S))}$, is a  Hilbert space isomorphism.
\end{theorem}
A kernel $k:S\times S\to\reals$ is called {\em normalized} if $\sup_{x\in S}k(x,x)=1$.
Also,
\begin{theorem}\label{thm:RKHS_kernel_from_basis}
Let $k:S\times S\to\mathbb R$ be a kernel and let $\{f_n\}_{n=1}^\infty$ be an orthonormal basis of a $H_k$.
Then, $k(x,y)=\sum_{n=1}^\infty f_n(x)f_n(y)$.
\end{theorem}

\subsubsection{Unitary Representations of Compact Groups}
Proofs of the results stated here can be found in \citep{Folland94}, chapter 5. Let $G$ be a compact group. A {\em unitary representation} (or just a {\em representation}) of $G$ is a group homomorphism $\rho:G\to U(H)$ where $U(H)$ is the class of unitary operators over a Hilbert space $H$, such that, for every $v\in H$, the mapping $a\mapsto \rho(a)v$ is continuous.

We say that a closed subspace $M\subset H$ is {\em invariant} (to $\rho$) if for every $a\in G, v\in M$, $\rho(a)v\in M$. We note that if $M$ is invariant then so is $M^\perp$.
We denote by $\rho|_{M}:G\to U(M)$ the {\em restriction} of $\rho$ to $M$. That is, $\forall a\in G,\;\rho|_M(a)=\rho(a)|_M$.
We say that $\rho:G\to U(H)$ is {\em reducible} if $H=M\oplus M^\perp$ such that $M,M^\perp$ are both non zero closed and invariant subspaces of $H$. A basic result is that every representation of a compact group is a sum of irreducible representation.

\begin{theorem}\label{thm:rep_decomposition}
Let $\rho:G\to U(H)$ be a representation of a compact group $G$. Then, $H=\oplus_{n\in I}H_n$, where every $H_n$ is invariant to $\rho$ and $\rho|_{H_n}$ is irreducible.
\end{theorem}
We shall also use the following Lemma.
\begin{lemma}\label{lem:rep_inuquness_on_inner_prod}
Let $G$ be a compact group, $V$ a finite dimensional vector space and let $\rho:G\to GL(V)$ be a continuous homomorphism of groups (here, $GL(V)$ is the group of invertible linear operators over $V$). Then,
\begin{enumerate}
\item There exists an inner product on $V$ making $\rho$ a unitary representation.
\item Moreover, if $V$ has no non-trivial invariant subspaces (here a subspace $U\subset V$ is called invariant if, $\forall a\in G,f\in U,\;\rho(a)f\in U$) then this inner product is unique up to scalar multiple.
\end{enumerate}
\end{lemma}

\subsubsection{Harmonic Analysis on the Sphere}
All the results stated here can be found in \citep{AtkinsonHa12}, chapters 1 and 2. Denote by $\mathbb O(d)$ the group of unitary operators over $\reals^d$ and by $dA$ the uniform probability measure over $\mathbb O(d)$ (that is, $dA$ is the unique probability measure satisfying $\int_{\mathbb O(d)}f(A)dA=\int_{\mathbb O(d)}f(AB) dA=\int_{\mathbb O(d)}f(BA) dA$ for every $B\in \mathbb O(d)$ and every integrable function $f:\mathbb O(d)\to \complex$).
Denote by $dx=dx_{d-1}$ the Lebesgue (area) measure over $S^{d-1}$ and let $L^2(S^{d-1}):=L^2(S^{d-1},dx)$. Given a measurable set $Z\subseteq S^{d-1}$, we sometime denote its Lebesgue measure by $|Z|$. Also, denote $dm=\frac{dx}{|S^{d-1}|}$ the Lebesgue measure, normalized to be a probability measure.

For every $n\in \mathbb N_0$, we denote by $\mathbb Y_n^d$ the linear space of $d$-variables harmonic (i.e., satisfying $\Delta p=0$) homogeneous polynomials of degree $n$. It holds that
\begin{equation}
N_{d,n}=\dim (\mathbb Y_n^d)=\binom{d+n-1}{d-1}-\binom{d+n-3}{d-1}=\frac{(2n+d-2)(n+d-3)!}{n!(d-2)!}
\end{equation}
Denote by $\mathcal P_{d,n}:L^2(S^{d-1})\to \mathbb Y_n^d$ the orthogonal projection onto $\mathbb Y_n^d$.

We denote by $\rho:\mathbb O(d)\to U(L^2(S^{d-1}))$ the unitary representation defined by
$$\rho(A)f=f\circ A^{-1}$$
We say that a closed subspace $M\subset L^2(S^{d-1})$ is {\em invariant} if it is invariant w.r.t. $\rho$ (that is, $\forall f\in M,A\in \mathbb O(d),\;f\circ A\in M$). We say that an invariant space $M$ is {\em primitive} if $\rho|_M$ is irreducible.

\begin{theorem}\label{thm:sherical_basic}
\begin{enumerate}
\item $L^2(S^{d-1})=\oplus_{n=0}^\infty\mathbb Y^d_n$.
\item The primitive finite dimensional subspaces of $L^2(S^{d-1})$ are exactly $\{\mathbb Y^d_n\}_{n=0}^\infty$.
\end{enumerate}
\end{theorem}

\begin{lemma}\label{lemma:spherical_addition}
Fix an orthonormal basis $Y^d_{n,j},\;1\le j\le N_{d,n}$ to $\mathbb Y^d_n$. For every $x\in S^{d-1}$ it holds that
$$\sum_{j=1}^{N_{d,n}}|Y^d_{n,j}(x)|^2=\frac{N_{d,n}}{|S^{d-1}|}$$
\end{lemma}

\subsubsection{Legendre and Chebyshev Polynomials}
The results stated here can be found at \citep{AtkinsonHa12}. Fix $d\ge 2$. The {\em $d$ dimensional Legendre polynomials} are the sequence of polynomials over $[-1,1]$ defined by the recursion formula
\begin{eqnarray*}
& P_{d,n}(x)=\frac{2n+d-4}{n+d-3}xP_{d,n-1}(x)+\frac{n-1}{n+d-3}P_{d,n-2}(x)\\
& P_{d,0}\equiv 1,\;P_{d,1}(x)=x
\end{eqnarray*}
We shall make use of the following properties of the Legendre polynomials.
\begin{proposition}\label{prop:Legendre}
\begin{enumerate}
\item For every $d\ge 2$, the sequence $\{P_{d,n}\}$ is orthogonal basis of the Hilbert space $L^2\left([-1,1],(1-x^2)^{\frac{d-3}{2}}dx\right)$.
\item For every $n,d$, $||P_{d,n}||_\infty=1$.
\end{enumerate}
\end{proposition}

The {\em Chebyshev polynomials of the first kind} are defined as $T_n:=P_{2,n}$. The {\em Chebyshev polynomials of the second kind} are the polynomials over $[-1,1]$ defined by the recursion formula
\begin{eqnarray*}
& U_n(x)=2xU_{n-1}(x)-U_{n-2}(x)\\
& U_0\equiv 1,\;U_1(x)=2x
\end{eqnarray*}
We shall make use of the following properties of the Chebyshev polynomials.
\begin{proposition}\label{prop:Chebyshev}
\begin{enumerate}
\item For every $n\ge 1$, $T_n'=nU_{n-1}$.
\item $||U_n||_{\infty}=n+1$.
\end{enumerate}
\end{proposition}
Given a measure $\mu$ over $[-1,1]$, {\em the orthogonal polynomials} corresponding to $\mu$ are the sequence of polynomials obtained upon the Gram-Schmidt procedure applied to $1,x,x^2,x^3,\ldots$. We note that the $1,\sqrt{2}T_1,\sqrt{2}T_2,\sqrt{2}T_3,\ldots$ are the orthogonal polynomials corresponding to the probability measure $d\mu = \frac{dx}{\pi\sqrt{1-x^2}}$

\subsubsection{Bochner Integral and Bochner Spaces}
Proofs and elaborations on the material appearing in this section can be found in \citep{Yosida63}. Let $(X,\mathfrak{m},\mu)$ be a measure space and let $H$ be a Hilbert space. A function $f:X\to H$ is {\em (Bochner) measurable} if there exits a sequence of function $f_n:X\to H$ such that
\begin{itemize}
\item For almost every $x\in X$, $f(x)=\lim_{n\to\infty}f_n(x)$.
\item The range of every $f_n$ is countable and, for every $v\in H$, $f^{-1}(v)$ is measurable.
\end{itemize}
A measurable function $f:X\to H$ is {\em (Bochner) integrable} if there exists a sequence of simple measurable functions (in the usual sense) $s_n$ such that $\lim_{n\to\infty}\int_X||f(x)-s_n(x)||_{H}d\mu(x)=0$. We define the integral of $f$ to be $\int_X fd\mu=\lim_{n\to\infty}\int s_nd\mu$, where the integral of a simple function $s=\sum_{i=1}^n 1_{A_i}v_i,A_i\in \mathfrak m,v_i\in H$ is $\int_Xsd\mu=\sum_{i=1}^n\mu(A_i)v_i$.

Define by $L^2(X,H)$ the Kolmogorov quotient (by equality almost everywhere) of all measurable functions $f:X\to H$ such that $\int_X||f||_H^2d\mu<\infty$. 
\begin{theorem}\label{thm:Bochner}
$L^2(X,H)$ in a Hilbert space w.r.t. the inner product $\langle f,g\rangle_{L^2(X,H)}=\int_{X}\langle f(x),g(x)\rangle_H d\mu(x)$
\end{theorem}

\subsection{Learnability implies small radius}\label{sec:C_is_small}
The purpose of this section is to show that if $\cx$ is a subset of some Hilbert space $H$ such that it is possible to learn affine functionals over $\cx$ w.r.t. some loss, then we can essentially assume that $\cx$ is contained is a unit ball and the returned affine functional is of norm $O\left(m^3\right)$, where $m$ is the number of examples.

\begin{lemma}[John's Lemma]\label{lem:John}\citep{Matousek02}
Let $V$ be an $m$-dimensional real vector space and let $K$ be a full-dimensional  compact convex set. There exists an inner product on $V$ so that $K$ is contained in a unit ball and contains a ball of radius $\frac{1}{m}$, both are centered at (the same) $x\in K$.
Moreover, if $K$ is $0$-symmetric it is possible to take $x=0$ and the ratio between the radiuses can be improved to $\sqrt{m}$.
\end{lemma}

\begin{lemma}\label{lem:fin_dim_to_inner_prod}
Let $l$ be a convex surrogate, let $V$ an $m$-dimensional vector space and let $\cx\subset V$ be a bounded subset that spans $V$ as an affine space. There exists an inner product $\inner{\cdot,\cdot}$  on $V$ and a probability measure $\mu_N$ such that
\begin{itemize}
\item For every $w\in V,b\in\mathbb R$, $||w||\le 4m^2\Err_{\mu_N,\hinge}(\Lambda_{w,b})$
\item $\cx$ is contained in a unit ball.
\end{itemize}
\end{lemma}
\proof
Let us apply John's Lemma
to $K=\operatorname{conv}(\cx)$. It yields an inner product on $V$ with $K$ contained in a unit ball and containing the ball with radius $\frac{1}{m}$ both centered at the same $x\in V$. 
It remains to prove the existence of the measure $\mu_N$. W.l.o.g., we assume that $x=0$.
 
Let $e_1,\ldots,e_m\in V$ be an orthonormal basis. For every $i\in [m]$, represent both $\frac{1}{m}e_i$ and $-\frac{1}{m}e_i$ as a convex combination of $m+1$ elements from $\cx$:
\[
\frac{1}{m}e_i=\sum_{j=1}^{m+1}\lambda_i^jx^j_i,\;\;-\frac{1}{m}e_i=\sum_{j=1}^{m+1}\rho_i^jz^j_i~.
\]
Now, define
\[
\mu_N(x^j_i,1)=\mu_N(x^j_i,-1)=\frac{\lambda_i^j}{4m},\;\;\mu_N(z^j_i,1)=\mu_N(z^j_i,-1)=\frac{\rho_i^j}{4m}~.
\] 
Finally, let $v\in V,b\in\mathbb R$. We have
\begin{eqnarray*}
\Err_{\mu_N,\hinge}(\Lambda_{w,b}) &=& \sum_{i=1}^m\sum_{j=1}^{m+1}\frac{\lambda_i^j}{4m}\left[l_{\hinge}(\Lambda_{w,b}(x_i^j))+l_{\hinge}(-\Lambda_{w,b}(x_i^j))\right]
\\
&&
+\frac{\rho_i^j}{4m}\left[l_{\hinge}(\Lambda_{w,b}(z_i^j))+l_{\hinge}(-\Lambda_{w,b}(z_i^j))\right]
\\
&\ge & \frac{1}{4m}\sum_{i=1}^m\left[l_{\hinge}\left(\sum_{j=1}^{m+1}\lambda_i^j(\Lambda_{w,b}(x_i^j))\right)
+l_{\hinge}\left(-\sum_{j=1}^{m+1}\lambda_i^j(\Lambda_{w,b}(x_i^j))\right)\right]
\\
&& +\left[l_{\hinge}\left(\sum_{j=1}^{m+1}\rho_i^j(\Lambda_{w,b}(z_i^j))\right)
+l_{\hinge}\left(-\sum_{j=1}^{m+1}\rho_i^j(\Lambda_{w,b}(z_i^j))\right)\right]
\\
&= & \frac{1}{4m}\sum_{i=1}^ml_{\hinge}\left(\left\langle w,\frac{e_i}{m}\right\rangle+b\right)
+l_{\hinge}\left(-\left\langle w,\frac{e_i}{m}\right\rangle-b\right)
\\
&&+l_{\hinge}\left(-\left\langle w,\frac{e_i}{m}\right\rangle+b\right)
+l_{\hinge}\left(\left\langle w,\frac{e_i}{m}\right\rangle-b\right)
\\
&\ge & \frac{1}{4m}\sum_{i=1}^ml_{\hinge}\left(-\frac{|\langle w,e_i\rangle|}{m}\right)
\\
&\ge & \frac{1}{4m^{2}}\sum_{i=1}^m|\langle w,e_i\rangle|
\\
&\ge & \frac{1}{4m^{2}}||w||
\end{eqnarray*}
\proofbox
Let $\cx$ be a bounded subset of some Hilbert space $H$ and let $C>0$.
Denote
\[
\cf_{H}(\cx,C)=\{\Lambda_{w,b}|_{\cx}\mid \|w\|_{H}\le C,b\in\mathbb R\}~.
\]
Denote by $\cm$ the collection of all affine subspaces of $H$ that are spanned by points from $\cx$. Denote by $t=t_{H}(\cx,C)$ be the maximal number such that for every affine subspace $M\in \cm$ of dimension less than $t$ there is $x\in \cx$ such that $d(x,M)>\frac{t}{C}$.
\begin{lemma}\label{lem:C_is_small_1}
Let $\cx$ be a bounded subset of some Hilbert space $H$ and let $C>0$. There is $A\subset \cx$ with $|A|=t_{H}(\cx,C)$ such that $[-1,1]^{A}\subset \cf_{H}(\cx,C)|_{A}$.
\end{lemma}
\proof
Denote $t=t_{H}(\cx,C)$. Let $x_0,\ldots,x_t\in\cx$ be points such that the ($t$ dimensional) volume of the parallelogram $Q$ defined by the vectors
\[
x_1-x_0,\ldots,x_t-x_0
\]
is maximal (if the supremum is not attained, the argument can be carried out with a parallelogram whose volume is sufficiently close to the supremum). Let $A=\{x_1,\ldots,x_t\}$.
We claim that for every $1\le i\le t$, the distance of $x_i$ from the affine span, $M_i$, of $A\cup\{x_0\}\setminus\{x_i\}$ is $\ge \frac{t}{C}$. Indeed, the volume of $Q$ is the ($t-1$ dimensional) volume of the parallelogram $Q\cap (M_i-x_0)$ times $d(x_i,M_i)$. By the maximality of $x_0,\ldots,x_t$ and the definition of $t$, $d(x_i,M_i)\ge \frac{t}{C}$.

For $1\le i\le t$ Let $v_i=x_i-P_{M_i}x_i$. Note that $\|v_i\|_{H}=d(x_i,M_i)\ge\frac{t}{C}$
Now, given a function $f:A\to [-1,1]$, we will show that $f\in \cf_{H}(\cx,C)$. Consider the affine functional
\[
\Lambda(x)=\sum_{i=1}^t \frac{f(x_i)}{\|v_i\|_H^2}\inner{v_i,x-P_{M_i}(x)}_H=\left\langle\sum_{i=1}^t\frac{f(x_i)v_i}{\|v_i\|_H^2},x\right\rangle_H-
\sum_{i=1}^t\left\langle\frac{f(x_i)v_i}{\|v_i\|_H^2},P_{M_i}(x)\right\rangle_H~.
\]
Note that since $v_i$ is perpendicular to $M_i$, $b:=-
\sum_{i=1}^t\left\langle\frac{f(x_i)v_i}{\|v_i\|_H^2},P_{M_i}(x)\right\rangle_H$ does not depend on $x$. Let $w:=\sum_{i=1}^t\frac{f(x_i)v_i}{\|v_i\|_H^2}$. We have
\[
\|w\|_{H}\le\sum_{i=1}^t |f(x_i)|\frac{1}{\|v_i\|}\le t\frac{C}{t}=C.
\]
Therefore, $\Lambda|_{\cx}\in\cf_{H}(\cx,C)$. Finally, for every $1\le j\le t$ we have
\[
\Lambda(x_j)=\sum_{i=1}^t\frac{f(x_i)}{\|v_i\|_H^2}\inner{v_i,v_j-P_i(v_j)}=f(x_i).
\]
Here, the last inequality follows form that fact that for $i\ne j$, since $v_i$ is perpendicular to $M_i$, $\inner{v_i,v_j-P_i(v_j)}=0$. Therefore, $f=\Lambda|_A$.
\proofbox
Let $l:\mathbb R\to\mathbb R$ be a surrogate loss function. We say that an algorithm 
{\em $(\epsilon,\delta)$-learns $\cf\subset \mathbb R^{\cx}$ using $m$ examples w.r.t. $l$} if:
\begin{itemize}
\item Its input is a sample of $m$ points in $\cx\times \{\pm 1\}$ and its output is a hypothesis in $\cf$.
\item For every  distribution $\cd$ on $\cx\times\{\pm 1\}$, it returns, with probability $1-\delta$, $\hat{f}\in\cf$ with $\Err_{\cd,l}(\hat{f})\le \inf_{f\in \cf}\Err_{\cd,l}(f)+\epsilon$
\end{itemize}

\begin{lemma}\label{lem:C_is_small_1.5}
Suppose that an algorithm $\ca$ $(\epsilon,\delta)$-learns $\cf$ using $m$ examples w.r.t. a surrogate loss $l$. Then, for every pair of distributions $\cd$ and $\cd'$ on $\cx\times \{\pm 1\}$, if $\hat{f}\in \cf$ is the hypothesis returned by $\ca$ running on $\cd$, then $\Err_{\cd',l}(f)\le m(l(0)+\epsilon)$ w.p. $\ge 1-2e\delta$.
\end{lemma}
\proof
Suppose toward a contradiction that w.p. $\ge 2e\delta$ we have $\Err_{\cd',l}(f)> a$ for $a> m(l(0)+\epsilon)$.
Consider the following distribution, $\tilde{\cd}$: w.p. $\frac{1}{m}$ we sample from $\cd'$ and with probability $1-\frac{1}{m}$ we sample from $\cd$.
Suppose now that we run the algorithm $\ca$ on $\tilde{\cd}$. 

Conditioning on the event that all the samples are from $\cd$, we have, w.p. $\ge 2e\delta$, $\Err_{\cd',l}(\hat{f})> a$ and therefore, $\Err_{\tilde{\cd},l}(\hat{f})> \frac{a}{m}$. The probability that indeed all the $m$ samples are from $\cd$ is $\left(1-\frac{1}{m}\right)^m>\frac{1}{2e}$. Hence, with probability $> \delta$, we have  $\Err_{\tilde{\cd},l}(\hat{f})> \frac{a}{m}$.

On the other hand, With probability $\ge 1-\delta$ we have $\Err_{\tilde{\cd},l}(\hat{f})\le \inf_{f\in\cf}\Err_{\tilde{\cd}}(f)+\epsilon\le \Err_{\cd,l}(0)+\epsilon= l(0)+\epsilon$. Hence, with positive probability,
\[
\frac{a}{m}\le l(0)+\epsilon~.
\]
It follows that $a\le m(l(0)+\epsilon)$.
\proofbox

\begin{lemma}\label{lem:C_is_small_main}
For every surrogate loss $l$ with $\partial_+l(0)<0$ there is a constant $c>0$ such that the following holds. 
Let $\cx$ be a bounded subset of a Hilbert space $H$.
If $\cf_{H}(\cx,C)$ is $(\epsilon,\delta)$-learnable using $m$ example w.r.t. $l$ then there is an inner product $\inner{\cdot,\cdot}_{s}$ on $H$ such that
\begin{itemize}
\item $\cx$ is contained in a unit ball w.r.t. $\|\cdot\|_s$.
\item If $\ca$ $(\epsilon,\delta)$-learns $\cf_H(\cx,C)$ then for every distribution $\cd$ on $\cx\times\{\pm 1\}$, the hypothesis $\Lambda_{w,b}$ returned by $\ca$ has $\|\Lambda_{w,b}\|_{s}\le c\cdot m^{3}$ w.p. $1-2e\delta$.
\item The norm $\|\cdot\|_{s}$ is equivalent\footnote{Two norms $\|\cdot\|$ and $\|\cdot\|'$ on a vector space $X$ are equivalent if for some $c_1,c_2>0$, $\forall x\in A,\;\;c_1\cdot\|x\|\le \|x\|'\le c_2\|x\|$} to $\|\cdot\|_{H}$.
\end{itemize}
\end{lemma}
\begin{remark}\label{rem:C_is_small_main}
If $\cx$ is the image of some mapping $\psi:\cz\to H$ then it follows from the lemma that there is a normalized kernel $k$ on $\cz$ such that the hypothesis returned by the learning algorithm (interpreted as a function from $\cz$ to $\mathbb R$) if the form $f+b$ with $\|f\|_k\le c\cdot m^3$. Also, if $\psi$ is continuous/absolutely continuous, then so is $k$.
\end{remark}

\proof
Let $t=t_H(\cx,C)$. By lemma \ref{lem:C_is_small_1} there is some $A\subset \cx$ such that $[-1,1]^{A}\subset \cf_H(\cx,C)|_A$. Since $\cf_H(\cx,C)$ is $(\epsilon,\delta)$-learnable using $m$ examples, it is not hard to see that we must have $t\le c'\cdot m$ for some $c'>0$ that depends only on $l$.

Therefore, there exists an affine subspace $M\subset H$ of dimension $d\le c'm$, such that for every $x\in\cx$, $d(x,M)<\frac{d}{C}\le \frac{c'm}{C}$. Moreover, we can assume that $M$ is spanned by some subset of $\cx$.
Denote by $\tilde{M}$ the linear space corresponding to to $M$ (i.e., $\tilde{M}$ is the translation of $M$ that contains $0$).
By lemma \ref{lem:fin_dim_to_inner_prod}, there is an inner product $\inner{\cdot,\cdot}_1$ on $\tilde{M}$, and a probability measure $\mu_{N}$ on $P_{\tilde{M}}\cx$ such that
\begin{itemize}
\item For every $w\in H,b\in \mathbb{R}$ we have $\|\Lambda_{P_{\tilde{M}}w,b}\|_1 \le 4d^{2}\Err_{\mu_N,\hinge}(\Lambda_{w,b})$. 
\item For all $x\in\cx$, $\|P_{\tilde{M}}x\|_1\le 1$.
\end{itemize}
Finally, define
\[
\inner{x,y}_{s}=\frac{1}{2}\inner{P_{\tilde{M}}(x),P_{\tilde{M}}(y)}_1+\frac{C^2}{2d^2}\inner{P_{\tilde{M}^{\perp}}(x),P_{\tilde{M}^{\perp}}(y)}_{H}~.
\]
The first and last assertions of the lemma are easy to verify, so we proceed to the second. Let $\Lambda_{w,b}$ be the hypothesis returned by $\ca$ after running on some $m$ examples sampled from some distribution $\cd$.
Let $\cd_N$ be a probability measure on $\cx$ whose projection on $\tilde{M}$ is $\mu_N$.
By lemma \ref{lem:C_is_small_1.5} we have, with probability $\ge 1-2e\delta$, $\Err_{\cd_N,l}(\Lambda_{w,b})\le (l(0)+1)m$. We claim that in this case $\Err_{\mu_N,\hinge}(\Lambda_{w,b})\le \left(\frac{l(0)+1}{\partial_+l(0)}+2\right)m$. Indeed,
\begin{eqnarray*}
\Err_{\mu_N,\hinge}(\Lambda_{w,b}) &=& \E_{(x,y)\sim\mu_N}l_{\hinge}(y\Lambda_{w,b}(x))
\\
&=& \E_{(x,y)\sim\cd_N}l_{\hinge}(y\Lambda_{w,b}(P_{\tilde{M}}x))
\\
&=& \E_{(x,y)\sim\cd_N}l_{\hinge}(y\Lambda_{w,b}(x+(x-P_{\tilde{M}}x)))
\\
&\le& \E_{(x,y)\sim\cd_N}l_{\hinge}(y\Lambda_{w,b}(x))+\|w\|_H\cdot d(x,P_{\tilde{M}}x)
\\
&\le& \Err_{\cd_N,\hinge}(\Lambda_{w,b})+C\cdot \frac{m}{C}
\\
&\le& \frac{1}{|\partial_+l(0)|} \Err_{\cd_N,l}(\Lambda_{w,b})+1+m\\
\\
&\le& \frac{l(0)+1}{\partial_+l(0)}m+2m
\end{eqnarray*}
By the properties of $\mu_N$, it follows that $\|\Lambda_{P_{\tilde{M}}w,b}\|_1=O\left( m^{3}\right)$ (here, the constant in the big-O notation depends only on $l$). Finally, we have,
\begin{eqnarray*}
\|\Lambda_w\|_{s}^2&=&\|\Lambda_{P_{\tilde{M}}w}\|_{s}^2+\|\Lambda_{P_{\tilde{M}^{\perp}}w}\|_{s}^2
\\
&=& 2\|\Lambda_{P_{\tilde{M}}w}\|_{1}^2+\frac{2m^2}{C^2}\|\Lambda_{P_{\tilde{M}^{\perp}}w}\|_{H}^2
\\
&\le& O\left(m^6\right)+2m^2\le O\left(m^6\right)
\end{eqnarray*}

\proofbox

\subsection{Symmetric Kernels and Symmetrization}
In this section we concern {\em symmetric kernels}. Fix $d\ge 2$ and let $k:S^{d-1}\times S^{d-1}\to\reals$ be a continuous positive definite kernel. We say that $k$ is {\em symmetric} if
$$\forall A\in\mathbb O(d),x,y\in S^{d-1},\;k(Ax,Ay)=k(x,y)$$
In other words, $k(x,y)$ depends only on $\langle x,y\rangle_{\reals^d}$. A RKHS is called {\em symmetric} if its reproducing kernel is symmetric. The next theorem characterize symmetric RKHSs. We note that Theorems of the same spirit have already been proved (e.g. \citep{Schoenberg1942}).

\begin{theorem}\label{thm:symmetric_kernels_characterization}
Let $k:S^{d-1}\times S^{d-1}\to \mathbb R$ be a normalized, symmetric and continuous kernel.
Then,
\begin{enumerate}
\item The group $\mathbb O(d)$ acts on $H_k$. That is, for every $A\in\mathbb O(d)$ and every $f\in H_k$ if holds that $f\circ A\in H_k$ and $||f||_{H_k}=||f\circ A||_{H_k}$.\label{symmetric_part_1}
\item The mapping $\rho:\mathbb O(d)\to U(H_k)$ defined by $\rho(A)f=f\circ A^{-1}$ is a unitary representation.\label{symmetric_part_2}
\item The space $H_k$ consists of continuous functions.\label{symmetric_part_2.5}

\item The decomposition of $\rho$ into a sum of irreducible representation is $H=\oplus_{n\in I}\mathbb Y^d_n$ for some set $I\subset \mathbb N_0$. Moreover,
$$\forall f,g\in H_k,\; \langle f,g\rangle_{H_k} =\sum_{n\in I}a^2_n\langle \mathcal P_{d,n}f,\mathcal P_{d,n}g\rangle_{L^2(S^{d-1})}$$
Where $\{a_n\}_{n\in I}$ are positive numbers.\label{symmetric_part_3}
\item It holds that $\sum_{n\in I} \frac{N_{d,n}}{|S^{d-1}|}a_n^{-2}=1$.\label{symmetric_part_4}
\end{enumerate}
\end{theorem}
\proof
Let $f\in H_k,\;A\in\mathbb O(d)$. To prove part \ref{symmetric_part_1}, assume first that
\begin{equation}\label{eq:2}
\forall x\in S^{d-1},\;f(x)=\sum_{i=1}^n\alpha_ik(x,y_i)
\end{equation}
For some $y_1,\ldots,y_n\in S^{d-1}$ and $\alpha_1,\ldots,\alpha_n\in\mathbb C$. We have, since $k$ is symmetric, that
\begin{eqnarray*}
f\circ A(x) &=& \sum_{i=1}^n \alpha_i k(Ax,y_i)\\
&=& \sum_{i=1}^n \alpha_i k(A^{-1}Ax,A^{-1}y_i)\\
&=& \sum_{i=1}^n \alpha_i k(x,A^{-1}y_i)
\end{eqnarray*}
Thus, by Theorem \ref{thm:RKHS_basic}, $f\circ A\in H_k$. Moreover, it holds that
\begin{eqnarray*}
||f\circ A||_{H_k}^2 &=& \sum_{1\le i,j\le n}\alpha_i\bar\alpha_j k(A^{-1}y_j,A^{-1}y_i)\\
&=& \sum_{1\le i,j\le n}\alpha_i\bar\alpha_j k(y_j,y_i)= ||f||_{H_k}^2
\end{eqnarray*}
Thus, part \ref{symmetric_part_1} holds for function $f\in H_k$ of the form (\ref{eq:2}). For general $f\in H_k$, by Theorem \ref{thm:RKHS_basic}, there is a sequence $f_n\in H_k$ of functions of the from (\ref{eq:2}) that converges to $f$ in $H_k$. From what we have shown for functions of the form (\ref{eq:2}) if follows that $||f_n-f_m||_{H_k}=||f_n\circ A-f_m\circ A||_{H_k}$, thus $f_n\circ A$ is a Cauchy sequence, hence, has a limit $g\in H_k$. By Theorem \ref{thm:RKHS_basic}, convergence in $H_k$ entails point wise convergence, thus, $g=f\circ A$. Finally,
$$||f||_{H_k}=\lim_{n\to\infty}||f_n||_{H_k}=\lim_{n\to\infty}||f_n\circ A||_{H_k}=||f\circ A||_{H_k}$$

We proceed to part \ref{symmetric_part_2}. It is not hard to check that $\rho$ is group homomorphism, so it only remains to validate that for every $f\in H$ the mapping $A\mapsto \rho(A)f$ is continuous. Let $\epsilon >0$ and let $A\in \mathbb O(d)$. We must show that there exists a neighbourhood $U$ of $A$ such that $\forall B\in U,\;||f\circ A^{-1}-f\circ B^{-1}||_{H_k}<\epsilon$. Choose $g(\cdot)=\sum_{i=1}^n\alpha_ik(\cdot,y_i)$ such that $||g-f||_{H_k}<\frac{\epsilon}{3}$. By part \ref{symmetric_part_1}, it holds that
\begin{eqnarray*}
||f\circ A^{-1}-f\circ B^{-1}||_{H_k} &\le & ||f\circ A^{-1}-g\circ A^{-1}||_{H_k}+||g\circ A^{-1}-g\circ B^{-1}||_{H_k}+||g\circ B^{-1}-f\circ B^{-1}||_{H_k}
\\
&=&||f-g||_{H_k}+||g\circ A^{-1}-g\circ B^{-1}||_{H_k}+||g-f||_{H_k}
\\
&<& \frac{\epsilon}{3}+||g\circ A^{-1}-g\circ B^{-1}||_{H_k}+\frac{\epsilon}{3}
\end{eqnarray*}
Thus, it is enough to find a neighbourhood $U$ of $A$ such that  $\forall B\in U,\;||g\circ A^{-1}-g\circ B^{-1}||_{H_k}<\frac{\epsilon}{3}$. However,
\begin{eqnarray*}
||g\circ A^{-1}-g\circ B^{-1}||_{H_k}^2 &=& ||g\circ A^{-1}||_{H_k}^2+||g\circ B^{-1}||_{H_k}^2-2\operatorname{Re}\left[\langle \sum_{i=1}^n\alpha_i k(\cdot,y_i)\circ A^{-1},\sum_{i=1}^n\alpha_i k(\cdot,y_i)\circ B^{-1}\rangle\right]
\\
&=& 2||g\circ A^{-1}||_{H_k}^2-2\operatorname{Re}\left[\langle \sum_{i=1}^n\alpha_i k(\cdot,A y_i),\sum_{i=1}^n\alpha_i k(\cdot,B y_i)\rangle\right]
\\
&=&  2||g\circ A^{-1}||_{H_k}^2-\operatorname{Re}\left[\sum_{i,j=1}^n\alpha_i\bar
\alpha_j k(By_j,A y_i)\right]
\end{eqnarray*}
Since $k$ is continuous, the last expression tends to $2||g\circ A^{-1}||_{H_k}^2-\operatorname{Re}\left[\sum_{i,j=1}^n\alpha_i\bar
\alpha_j k(Ay_j,A y_i)\right]=||g\circ A-g\circ A||_{H_k}^2=0$ as $B\to A$. Thus, there exists a neighbourhood $U$ such that $\forall B\in U,\;||g\circ A^{-1}-g\circ B^{-1}||_{H_k}<\frac{\epsilon}{3}$ as required.

To see part \ref{symmetric_part_2.5}, note that every function in $H_k$ is a limit in $H_k$ of functions of the form (\ref{eq:2}). Since $k$ is continuous, every function in $H_k$ is a limit in $H_k$ of continuous functions. However, by Theorem \ref{thm:RKHS_basic}, every function is in fact a uniform limit of continuous function, thus -- continuous itself.

We proceed to part \ref{symmetric_part_3}. By Theorem \ref{thm:rep_decomposition} $H_k=\oplus_{i\in I}V_i$ where each $V_i$ is a finite dimensional space that is invariant to $\rho$. By Theorem \ref{thm:sherical_basic} each $V_i$ must be $Y_n$ for some $n$, thus, $H=\oplus_{n\in I}\mathbb Y^d_n$. By the uniqueness part in Lemma \ref{lem:rep_inuquness_on_inner_prod} and Theorem \ref{thm:sherical_basic}, the restriction of $\langle\cdot,\cdot\rangle_{H_k}$ to each $\mathbb Y^d_n,\;n\in I$ equals to  $\langle\cdot,\cdot\rangle_{L^2(S^{d-1})}$ up to scalar multiple, proving the formula for $\langle\cdot,\cdot\rangle_{H_k}$

Finally, to see equation part \ref{symmetric_part_4}, note that if for every $n\in I$, $\{Y^d_{n,j}\}_{j\in [N_{d,n}]}$ in an orthonormal basis of $\mathbb Y^d_n$ w.r.t. $\langle \cdot,\cdot\rangle_{L^2(S^{d-1})}$ then $\{\frac{1}{a_n}Y^d_{n,j}\}_{n\in I,j\in [N_{d,n}]}$ is an orthogonal basis of $H$. By Theorem \ref{thm:RKHS_kernel_from_basis} and Lemma \ref{lemma:spherical_addition}, it follows that, for every $x \in S^{d-1}$,
$$1=k(x,x)=\sum_{n\in I} a_n^{-2}\sum_{j=1}^{N_{d,n}} (Y^d_{n,j}(x))^2=\sum_{n\in I} \frac{N_{d,n}}{|S^{d-1}|}a_n^{-2}$$
\proofbox

\subsubsection*{Symmetrization}
Let $k:S^{d-1}\times S^{d-1}\to \reals$ be a normalized continuous kernel. We define its {\em symmetrization} by
$$\forall x,y\in S^{d-1},\;k_s(x,y)=\int_{\mathbb O(d)}k(Ax,Ay)dA$$
\begin{theorem}\label{thm:symmetrization}
\begin{enumerate}
\item $k_s$ is symmetric continuous kernel with $\sup_{x\in S^{d-1}}k_s(x,x)\le 1$.
\item For every $\Phi\in L^2(\mathbb O(d),H_k)$ define $\bar \Phi:S^{d-1}\to \mathbb C$ by $\bar \Phi(x)=\int_{\mathbb O(d)}\Phi(A)(Ax)dA$. Then
$$H_{k_s}=\{\bar \Phi:\Phi\in L^2(\mathbb O(d),H_k)\}$$
Moreover, for every $\Phi\in L^2(\mathbb O(d),H_k)$, $||\bar\Phi||_{H_{k_s}}\le ||\Phi||_{L^2(\mathbb O(d),H_k)}$.
\end{enumerate}
\end{theorem}
\proof
Part 1. follows readily from the definition. We proceed to part 2. Define $\phi:S^{d-1}\to L^2(\mathbb O(d),H_k)$ by 
$$\phi(x)(A)(\cdot)=k(Ax,\cdot)$$
Note that
\begin{eqnarray*}
\langle \phi(x),\phi(y)\rangle_{L^2(\mathbb O(d),H_k)} &=& \int_{\mathbb O(d)}\langle\phi(x)(A),\phi(y)(A)\rangle\\
&=&\int_{\mathbb O(d)}\langle\phi(x)(A),\phi(y)(A)\rangle\\
&=& k_s(x,y)
\end{eqnarray*}
Thus, the Theorem follows from Theorem \ref{thm:RKHS_embedding}
\proofbox

\subsection{Lemma~\ref{lemma:changes_slowly_high_dim} and its proof}
\begin{lemma}\label{lemma:Legendre_bound}
For every $n>0,d\ge 5$ and $t\in [-1,1]$ it holds that
$$|P_{d,n}(t)|\le \min\left\{\frac{\Gamma\left(\frac{d-1}{2}\right)}{\sqrt{\pi}}\left[\frac{4}{n(1-t^2)}\right]^{\frac{d-2}{2}},\left(\frac{n}{n+d-2}+2|t|\right)^\frac{n}{2}\right\}$$
Moreover, if $\frac{n}{n+d-2}+2|t|\le 1$ we also have
$$|P_{d,n}(t)|\le \sqrt{\prod_{i=1}^n\left(\frac{i}{i+d-2}+2|t|\right)}$$
Finally, there exist constants $E>0$ and $0<r,s<1$ such that for every $K>0,d\ge 5$ and $t\in \left[-\frac{1}{8},\frac{1}{8}\right]$ we have
$$\sum_{n=K}^\infty |P_{d,n}(t)| \le Er^K+Es^d$$

\end{lemma}
\proof
In \citep{AtkinsonHa12} it is shown that $|P_{d,n}(t)|\le \frac{\Gamma\left(\frac{d-1}{2}\right)}{\sqrt{\pi}}\left[\frac{4}{n(1-t^2)}\right]^{\frac{d-2}{2}}$.
We shall prove, by induction on $k$ that
$$|P_{d,n}(t)|\le \sqrt{\prod_{i=1}^n\left(\frac{i}{i+d-2}+2|t|\right)}$$
Whenever $\frac{n}{n+d-2}+2|t|\le 1$. For $n=0,1$ it follows from the fact that $P_{d,0}\equiv 1$ and $P_{d,1}(t)=t$. Let $n>1$. By the induction hypothesis and the recursion formula for the Legendre polynomials we have 
\begin{eqnarray*}
|P_{d,n}(t)| &\le& \frac{2n+d-4}{n+d-3}|t||P_{d,n-1}(t)|+\frac{n-1}{n+d-3}|P_{d,n-2}(t)|
\\
&\le &2|t||P_{d,n-1}(t)|+\frac{n-1}{n+d-3}|P_{d,n-2}(t)|
\\
&\le &2|t|\sqrt{\prod_{i=1}^{n-1}\left(\frac{i}{i+d-2}+2|t|\right)}+\frac{n-1}{n+d-3}\sqrt{\prod_{i=1}^{n-2}\left(\frac{i}{i+d-2}+2|t|\right)}
\\
&\le &2|t|\sqrt{\prod_{i=1}^{n-2}\left(\frac{i}{i+d-2}+2|t|\right)}+\frac{n-1}{n+d-3}\sqrt{\prod_{i=1}^{n-2}\left(\frac{i}{i+d-2}+2|t|\right)}
\\
&\le &\sqrt{\left(2|t|+\frac{n-1}{n+d-3}\right)\left(2|t|+\frac{n}{n+d-2}\right)}\sqrt{\prod_{i=1}^{n-2}\left(\frac{i}{i+d-2}+2|t|\right)}
\\
&=& \sqrt{\prod_{i=1}^{n}\left(\frac{i}{i+d-2}+2|t|\right)}
\end{eqnarray*}

Now, every $K,\bar K\ge 0$ such that $\left(\frac{\bar K}{\bar K+d-2}+2|t|\right)^\frac{1}{2}<1$, we have
\begin{eqnarray*}
\sum_{n=K}^\infty |P_{d,n}(t)| &\le & \sum_{n=K}^{\bar K}\left(\frac{n}{n+d-2}+2|t|\right)^\frac{n}{2}+\sum_{n=\bar K+1}^\infty\frac{\Gamma\left(\frac{d-1}{2}\right)}{\sqrt{\pi}}\left[\frac{4}{n(1-t^2)}\right]^{\frac{d-2}{2}}
\\
&\le &\sum_{n=K}^{\bar K}\left(\frac{\bar K}{\bar K+d-2}+2|t|\right)^\frac{n}{2}+\sum_{n=\bar K+1}^\infty\frac{\Gamma\left(\frac{d-1}{2}\right)}{\sqrt{\pi}}\left[\frac{4}{n(1-t^2)}\right]^{\frac{d-2}{2}}
\\
&\le &\sum_{n=K}^{\infty}\left(\frac{\bar K}{\bar K+d-2}+2|t|\right)^\frac{n}{2}+\frac{\Gamma\left(\frac{d-1}{2}\right)}{\sqrt{\pi}}\left[\frac{4}{(1-t^2)}\right]^{\frac{d-2}{2}}\sum_{n=\bar K+1}^\infty n^{-\frac{d-2}{2}}
\\
&\le &\frac{\left(\frac{\bar K}{\bar K+d-2}+2|t|\right)^\frac{K}{2}}{1-\left(\frac{\bar K}{\bar K+d-2}+2|t|\right)^\frac{1}{2}}+\frac{\Gamma\left(\frac{d-1}{2}\right)}{\sqrt{\pi}}\left[\frac{4}{(1-t^2)}\right]^{\frac{d-2}{2}}\sum_{n=\bar K+1}^\infty n^{-\frac{d-2}{2}}
\\
&\le &\frac{\left(\frac{\bar K}{\bar K+d-2}+2|t|\right)^\frac{K}{2}}{1-\left(\frac{\bar K}{\bar K+d-2}+2|t|\right)^\frac{1}{2}}+\frac{\Gamma\left(\frac{d-1}{2}\right)}{\sqrt{\pi}}\left[\frac{4}{(1-t^2)}\right]^{\frac{d-2}{2}}\int_{\bar K}^\infty x^{-\frac{d-2}{2}}dx
\\
&= &\frac{\left(\frac{\bar K}{\bar K+d-2}+2|t|\right)^\frac{K}{2}}{1-\left(\frac{\bar K}{\bar K+d-2}+2|t|\right)^\frac{1}{2}}+\frac{\Gamma\left(\frac{d-1}{2}\right)}{\sqrt{\pi}}\left[\frac{4}{(1-t^2)}\right]^{\frac{d-2}{2}}\frac{{\bar K}^{-\frac{d-4}{2}}}{\frac{d-4}{2}}
\end{eqnarray*}
(We limit ourselves to $d \ge 5$ to guarantee the convergence of $\sum n^{-\frac{d-2}2}$.)
In particular, if $|t|\le \frac{1}{8}$ and $\bar K=d-2$, we have,
\begin{eqnarray*}
\sum_{n=K}^\infty |P_{d,n}(t)| &\le & \left(\frac{1}{1-\left(\frac{3}{4}\right)^{\frac{1}{2}}}\right)\left(\frac{3}{4}\right)^{\frac{K}{2}}+\frac{\Gamma\left(\frac{d-1}{2}\right)}{\sqrt{\pi}}\left[\frac{4.07}{(d-2)}\right]^{\frac{d-2}{2}}\frac{d-2}{\frac{d-4}{2}}
\\
&\le & \left(\frac{1}{1-\left(\frac{3}{4}\right)^{\frac{1}{2}}}\right)\left(\frac{3}{4}\right)^{\frac{K}{2}}+\frac{6\Gamma\left(\frac{d-1}{2}\right)}{\sqrt{\pi}}\left[\frac{4.07}{(d-2)}\right]^{\frac{d-2}{2}}
\\
&\sim & \left(\frac{1}{1-\left(\frac{3}{4}\right)^{\frac{1}{2}}}\right)\left(\frac{3}{4}\right)^{\frac{K}{2}}+\frac{6}{\sqrt{\pi}}\left[\frac{4.07}{(d-2)}\right]^{\frac{d-2}{2}}\sqrt{\frac{2\pi}{\frac{d-2}{2}}}\left(\frac{d-2}{2e}\right)^{\frac{d-2}{2}}
\\
&= & \left(\frac{1}{1-\left(\frac{3}{4}\right)^{\frac{1}{2}}}\right)\left(\frac{3}{4}\right)^{\frac{K}{2}}+12\left[\frac{4.07}{2e}\right]^{\frac{d-2}{2}}
\end{eqnarray*}
\proofbox
\begin{lemma}\label{lemma:l_1-l_2}
Let $\mu$ be a probability measure on $[-1,1]$ and let $p_0,p_1,\ldots$ be the corresponding orthogonal polynomials. 
Then, for every $f\in\operatorname{span} \{p_0,\ldots,p_{K-1}\}$ we have
$$||f||_2\le \sqrt{K}||f||_1\cdot\max_{0\le i\le K-1}||p_i||_{\infty}$$
Here, all $L^p$ norms are w.r.t. $\mu$.
\end{lemma}
\proof
Write $f=\sum_{i=0}^{K-1}\alpha_ip_i$ and denote $M=\max_{0\le i\le K-1}||p_i||_{\infty}. $We have
\begin{eqnarray*}
|| f||_2^2 &\le & || f||_1\cdot|| f||_\infty
\\
&\le & ||f||_1\cdot M\sum_{n=0}^{K-1}|\alpha_k|
\\
&\le & ||f||_1\cdot M\sqrt{\sum_{n=0}^{K-1} \alpha_k^2} \cdot\sqrt{K}
\\
&=& ||f||_1\cdot M\cdot || f||_2 \sqrt{K}
\end{eqnarray*}
\proofbox

\begin{lemma}\label{lemma:changes_slowly}
Let $d\ge 5$ and let $f:[-1,1]\to \reals$ be a continuous function whose expansion in the basis of $d$-dimensional Legendre polynomials is
$$f=\sum_{n=0}^\infty \alpha_nP_{d,n}$$
Denote $C=\sup_{n} |\alpha_n|$. Let $\mu$ be the probability measure on $[-1,1]$ whose density function is
$$w(x)=\begin{cases}
0 & |x|>\frac{1}{8}\\
\frac{8}{\pi\sqrt{1-\left( 8x\right)^2}} & |x|\le \frac{1}{8}
\end{cases}$$
Then, for every $K\in\mathbb N,\frac{1}{8}>\gamma>0$,
$$|f(\gamma)-f(-\gamma)|\le  32 \gamma K^{3.5} \cdot  ||f||_{1,\mu}+\left(32 \gamma K^{3.5}+2\right)\cdot C\cdot E\cdot (r^K+s^d)$$
Here, $E,r$ and $s$ are the constants from Lemma \ref{lemma:Legendre_bound}.
\end{lemma}
\proof
Let $\bar f=\sum_{n=0}^{K-1}\alpha_nP_{d,n}$. We have $||\bar f||_{1,\mu}\le ||f||_{1,\mu}+||\bar f-f||_{\infty,\mu}$. Define $g:[-1,1]\to\reals$ by $g(x)=\bar f (\frac{x}{8})$ and denote by $d\lambda=\frac{dx}{\pi\sqrt{1-x^2}}$. Write,
$$g=\sum_{n=0}^{K-1}\beta_nT_n$$
Where $T_n$ are the Chebyshev polynomials. By Lemma \ref{lemma:l_1-l_2} it holds that, for every $0\le n\le K-1$,
$$|\beta_n|\le \sqrt{2}||g||_{2,\lambda}\le 2\sqrt{K}||g||_{1,\lambda}=2\sqrt{K}||\bar f||_{1,\mu}$$
Now,
$$g'=\sum_{n=1}^{K-1} \beta_knU_{n-1}$$
Where $U_n$ are the Chebyshev polynomials of the second kind.
Thus,
$$||g'||_{\infty,\lambda}\le \sum_{n=1}^{K-1} |\beta_k|\cdot n\cdot||U_{n-1}||_{\infty,\lambda} = \sum_{n=1}^{K-1} |\beta_k|\cdot n^2\le 2\sqrt{K}||\bar f||_{1,\mu}\cdot K^3$$
Finally, by Lemma \ref{lemma:Legendre_bound},
\begin{eqnarray*}
|f(\gamma)-f(-\gamma)| &\le & |g(8\gamma)-g(-8\gamma)|+2||f-\bar f||_{\infty,\mu}
\\
&\le & 32 \gamma K^{3.5} \cdot ||\bar f||_{1,\mu}+2||f-\bar f||_{\infty,\mu}
\\
&\le & 32 \gamma K^{3.5} \cdot \left( ||f||_{1,\mu}+||f-\bar f||_{\infty,\mu} \right)+2||f-\bar f||_{\infty,\mu}
\\
&\le & 32 \gamma K^{3.5} \cdot  ||f||_{1,\mu}+\left(32 \gamma K^{3.5}+2\right)\cdot ||f-\bar f||_{\infty,\mu}
\\
&\le & 32 \gamma K^{3.5} \cdot  ||f||_{1,\mu}+\left(32 \gamma K^{3.5}+2\right)\cdot E\cdot C\cdot (r^K+s^d)
\end{eqnarray*}
\proofbox

For $e\in S^{d-1}$ we define the group $\mathbb O(e):=\{A\in\mathbb O(d):Ae=e\}$.
If $H_k$ be a symmetric RKHS and $e\in S^{d-1}$ we define {\em Symmetrization around $e$}. This is the operator $\mathcal P_e:H_k\to H_k$ which is the projection on the subspace $\{f\in H_k:\forall A\in \mathbb O(e),\;f\circ A=f\}$. It is 
not hard to see that $(\mathcal P_ef)(x)=\int_{\{x':\langle x',e\rangle=\langle x,e\rangle\}}f(x')dx'=\int_{\mathbb O(e)}f\circ A(x)dA$. Since $\mathcal P_ef$ is a convex combination of the functions $\{f\circ A\}_{A\in \mathbb O(e)}$, it follows that if $\mathcal R:H_k\to \reals$ is a convex functional then $\mathcal R(\mathcal P_ef)\le \int_{\mathbb O(e)}\mathcal R(f\circ A)dA$.

\begin{lemma}[main]\label{lemma:changes_slowly_high_dim}
There exists a probability measure $\mu$ on $[-1,1]$ with the following properties.
For every continuous and normalized kernel $k:S^{d-1}\times S^{d-1}\to \reals$ and $C>0$, there exists $e\in S^{d-1}$ such that, for every $f\in H_k$ with $||f||_{H_k}\le C$, $K\in\mathbb N$ and $0<\gamma<\frac{1}{8}$,
\begin{eqnarray*}
\left| \int_{\{x:\langle x,e\rangle=\gamma\}}f-\int_{\{x:\langle x,e\rangle=-\gamma\}}f\right| &\le &32 \gamma K^{3.5} \cdot  ||f||_{1,\mu_e}+\left(32 \gamma K^{3.5}+2\right)\cdot E\cdot C\cdot (r^K+s^d)
\\
&\le & 32 \gamma K^{3.5} \cdot  ||f||_{1,\mu_e}+10\cdot E\cdot K^{3.5} \cdot C\cdot (r^K+s^d)
\end{eqnarray*}
The integrals are w.r.t. the uniform probability over $\{x:\langle x,e\rangle=\gamma\}$ and $\{x:\langle x,e\rangle=-\gamma\}$ and $E,r,s$ are the constants from Lemma \ref{lemma:Legendre_bound}.
\end{lemma}
\proof
Suppose first that $k$ is symmetric. Let $\mu$ be the distribution over $[-1,1]$ whose density function is

$$w(x)=\begin{cases}
0 & |x|>\frac{1}{8}\\
\frac{8}{\pi\sqrt{1-\left( 8x\right)^2}} & |x|\le \frac{1}{8}
\end{cases}$$
We can assume that $f$ is $\mathbb O(e)$-invariant. Otherwise, we can replace $f$ with $\mathcal P_ef$, which does not change the l.h.s. and does not increase the r.h.s. This assumption yields (see \citep{AtkinsonHa12}, pages 17-18)
$$f(x)=\sum_{n=0}^\infty \alpha_n P_{d,n}(\langle e,x\rangle).$$
The $L^2(S^{d-1})$-norm of the map
$x\mapsto P_{d,n}(\langle x,e\rangle)$ is $\frac{|S^{d-1}|}{N_{d,n}}$ (e.g. \citep{AtkinsonHa12}, page 71). Therefore,
$$||f||_k^2=\sum_{n\in I} \frac{|S^{d-1}|}{N_{d,n}}a_n^2\alpha_n^2$$
where $\{a_n\}_{n\in I}$ are the numbers corresponding to $H_k$ from Theorem \ref{thm:symmetric_kernels_characterization}.
In particular (since also for $n\not\in I$, $\alpha_n=0$),
$$|\alpha_n|^2\le \frac{N_{d,n}}{|S^{d-1}|}a_n^{-2}||f||_k^2\le ||f||_k^2$$
Write
$$g(t)=f(te),\;t\in [-1,1]$$
By Lemma \ref{lemma:changes_slowly},
$$|g(\gamma)-g(-\gamma)|\le  32 \gamma K^{3.5} \cdot  ||f||_{1,\mu}+\left(32 \gamma K^{3.5}+2\right)\cdot E\cdot C\cdot (r^K+s^d)$$
Finally, $\int_{\{x:\langle x,e\rangle=\gamma\}}f=g(\gamma),\;\int_{\{x:\langle x,e\rangle=-\gamma\}}f=g(-\gamma)$ since $f$ is $\mathbb O(e)$-invariant. The Lemma follows.

We proceed to the general case where $k$ is not necessarily symmetric. Assume by way of contradiction that for every $e\in S^{d-1}$, there  exists a function $f_e$ such that 
\begin{equation}\label{eq:4}
\int_{\{x:\langle x,e\rangle=\gamma\}}f_e-\int_{\{x:\langle x,e\rangle=-\gamma\}}f_e > 32 \gamma K^{3.5} \cdot  ||f_e||_{1,\mu_e}+\left(32 \gamma K^{3.5}+2\right)\cdot ||f_e||_{H_k}\cdot C\cdot (r^K+s^d)
\end{equation}
For convenience we normalize, so l.h.s. equals $1$. Fix a vector $e_0\in S^{d-1}$. Define $\Phi\in L^2(\mathbb O(d),H_k)$ by
$$\Phi(A)=f_{Ae_0}$$
and let $f\in H_{k_s}$ be the function
$$f(x)=\int_{\mathbb O(d)}\Phi(A)(Ax)dA=\int_{\mathbb O(d)}f_{Ae_0}(Ax)dA$$
Now, it holds that
\begin{eqnarray*}
\int_{\{x:\langle x,e_0\rangle=\gamma\}}f-\int_{\{x:\langle x,e_0\rangle=-\gamma\}}f &=&\int_{\{x:\langle x,e_0\rangle=\gamma\}}\int_{\mathbb O(d)}f_{Ae_0}(Ax)dAdx-\int_{\{x:\langle x,e_0\rangle=-\gamma\}}\int_{\mathbb O(d)}f_{Ae_0}(Ax)dAdx
\\
&=&\int_{\mathbb O(d)}\int_{\{x:\langle x,e_0\rangle=\gamma\}}f_{Ae_0}(Ax)dx-\int_{\{x:\langle x,e_0\rangle=-\gamma\}}f_{Ae_0}(Ax)dxdA\\
\\
&=&\int_{\mathbb O(d)}\int_{\{x:\langle x,Ae_0\rangle=\gamma\}}f_{Ae_0}(x)dx-\int_{\{x:\langle x,Ae_0\rangle=-\gamma\}}f_{Ae_0}(x)dxdA\\
&=& 1
\end{eqnarray*}
On the other hand 
\begin{eqnarray*}
||f||_{1,\mu_e}&=& \int_{S^{d-1}}\left|\int_{\mathbb O(d)}f_{Ae_0}(Ax)dA\right|d\mu_{e_0}(x)
\\
&\le & \int_{\mathbb O(d)}\int_{S^{d-1}}\left| f_{Ae_0}(Ax)\right|d\mu_{e_0}(x)dA
\\
&\le & \int_{\mathbb O(d)}\int_{S^{d-1}}\left| f_{Ae_0}(x)\right|d\mu_{A{e_0}}(x)dA
\\
&=& \int_{\mathbb O(d)}|| f_{Ae_0}||_{1,\mu_{Ae_0}}dA
\end{eqnarray*}
Moreover, by Theorem \ref{thm:symmetrization},
$$||f||^2_{H_{k_s}}\le ||\Phi ||_{L^2(\mathbb O(d),H_k)}^2=\int_{\mathbb O(d)}||f_{Ae_0}||_{H_k}^2dA\le C^2$$
Since the Lemma is already proved for symmetric kernels, it follows that
\begin{eqnarray*}
1 &\le & 32 \gamma K^{3.5} \cdot  ||f||_{1,\mu_{e_0}}+\left(32 \gamma K^{3.5}+2\right)\cdot E\cdot C\cdot (r^K+s^d)
\\
&\le & 32 \gamma K^{3.5} \cdot  \int_{\mathbb O(d)}|| f_{Ae_0}||_{1,\mu_{Ae_0}}dA+\left(32 \gamma K^{3.5}+2\right)\cdot E\cdot C\cdot (r^K+s^d)
\\
&= & \int_{\mathbb O(d)} 32 \gamma K^{3.5} \cdot || f_{Ae_0}||_{1,\mu_{Ae_0}}+\left(32 \gamma K^{3.5}+2\right)\cdot E\cdot C\cdot (r^K+s^d)dA
\end{eqnarray*}
Thus, for some $A\in \mathbb O(d)$
$$1\le 32 \gamma K^{3.5} \cdot || f_{Ae_0}||_{1,\mu_{Ae_0}}+\left(32 \gamma K^{3.5}+2\right)\cdot E\cdot C\cdot (r^K+s^d)$$
Contradicting Equation (\ref{eq:4}).
\proofbox

\subsection{Proofs of the main Theorems}
We are now ready to prove Theorems \ref{thm:main_1} and \ref{thm:main_3}. We only consider distributions that supported on the unit sphere, and we can therefore assume that the problem is formulated it terms of the unit sphere and not the unit ball. Also, we reformulate program (\ref{eq:svm_1_general}) as follows: Given $l:\reals\to\reals$ a convex surrogate, a constant $C>0$ and a continuous kernel $k:S^\infty\times S^\infty\to \reals$ with $\sup_{x\in S^\infty}k(x,x)\le 1$, we want to solve
\begin{eqnarray}\label{eq:svm_equiv}
\min &&  \Err_{\mathcal D,l}\left(f+b\right) \nonumber \\
\text{s.t.} && f\in H_k,\;b\in\reals\\
&& ||f||_{H_k}\le C\nonumber
\end{eqnarray}
We can assume that $\partial_+l(0)<0$, for otherwise the approximation ratio is $\infty$. To see that, let the distribution $\mathcal D$ be concentrated on a single point on the sphere and always return the label $1$. Of course, $\Err_\gamma(\mathcal D)=0$. However, if $\partial_+l(0)\ge 0$, it is bot hard to see that if $f,b$ is the solution of program (\ref{eq:svm_equiv}), then $f(x)+b\le 0$, so that $\Err_{0-1}(f+b)=1$.

\begin{lemma}\label{lemma:L1_err}
Let $l$ be a surrogate loss, $\mu$ a probability measure on $S^{d-1}$ and $f\in C(S^{d-1})$. Let $\bar\mu$ be the probability measure on $S^{d-1}\times \{\pm 1\}$ which is the product measure of $\mu$ and the uniform distribution on $\{\pm 1\}$. Then
$$||f||_{1,\mu}\le \frac{2}{|\partial_+l(0)|}\Err_{\bar\mu,l}(f)$$
\end{lemma}
\proof 
By Jansen's inequaliy, it holds that
\begin{eqnarray*}
\Err_{\bar \mu ,l}(f) &= & \mathbb E_{(x,y)\sim \bar \mu}l(y\cdot f(x))
\\
&= & \frac{1}{2}  \mathbb E_{(x,y)\sim  \bar \mu} l(f(x))+l(-f(x))
\\
&\ge & \frac{1}{2} \mathbb E_{(x,y)\sim \bar \mu} l(-|f(x)|)
\\
&\ge & \frac{1}{2}l\left(-\mathbb E_{(x,y)\sim  \bar\mu} |f(x)|\right)
\end{eqnarray*}
It follows that $l\left(-||f||_{1,\mu} \right)\le 2 \Err_{\bar\mu , l}(f)$. By the convexity of $l$, it follows that for every $x\in \reals$, $l(x)\ge l(0)+x\cdot\partial_+l(0)= l(0)-x\cdot|\partial_+l(0)|\ge -x\cdot|\partial_+l(0)|$. Thus,
$$||f||_{1,\mu}\le \frac{2}{|\partial_+l(0)|} \Err_{\bar\mu , l}(f)$$
\proofbox

\subsubsection{Theorems \ref{thm:main_1} and \ref{thm:main_3}}
We will need Levy's measure concentration Lemma (e.g., \citep{MilmanSc02}). Let $f:X\to Y$ be an absolutely continuous map between metric spaces. We define its {\em modulus of continuity} as
$$\forall \epsilon>0,\;\omega_f(\epsilon)=\sup\{d(f(x),f(y)):x,y\in X,d(x,y)\le\epsilon\}$$

\begin{theorem}[Levy's Lemma]\label{thm:Levi}
There exists a constant $\eta>0$ such that for every continuous function $f:S^{d-1}\to \reals$,
$$\Pr\left(|f-\mathbb Ef|>\omega_f(\epsilon)\right)\le \exp\left(-\eta d\epsilon^2\right)$$
Here, both probability and expectation are w.r.t. the uniform distribution.
\end{theorem}
We note that $\omega_{f\circ g}\le\omega_{f}\cdot \omega_{g}$ and that $\omega_{\Lambda_v}(\epsilon)=\|v\|\cdot\epsilon$. Thus, if $\psi:S^{\infty}\to H_1$ is an absolutely continuous embedding such that $k(x,y)=\langle \psi(x),\psi(y)\rangle_{H_1}$, then 
for every $v\in H_1$, it holds that $\omega_{\Lambda_{v,0}\circ \psi}\le ||v||_{H_1}\cdot \omega_{\psi}$. 
Suppose now that $f\in H_k$ with $\|f\|_{H_k}\le C$. Let $v\in H_1$ such that $f=\Lambda_{v,0}\circ\psi$ and $||v||_{H_1}=||f||_{H_k}\le C$. It follows from Levi's Lemma that
\begin{equation}\label{eq:Levi}
\Pr\left(|f- \mathbb Ef|>C\cdot \omega_{\psi}(\epsilon)\right)\le\Pr\left(|f-\mathbb Ef|>\omega_f(\epsilon)\right)\le \exp\left(-\eta d\epsilon^2\right)
\end{equation}
Again, when both probability and expectation are w.r.t. the uniform distribution over $S^{d-1}$.

\proof (of Theorems \ref{thm:main_1} and \ref{thm:main_3})
Let $\beta>\alpha>0$ such that $l(\alpha)>l(\beta)$. Choose $0<\theta<1$ large enough so that $(1-\theta)l(-\beta)+\theta l(\beta)<\theta l(\alpha)$.
Define probability measures $\mu^1,\mu^2,\mu$ over $[-1,1]\times\{\pm 1\}$ as follows. 
$$\mu^1((-\gamma,-1))=1-\theta,\;\mu^1((\gamma,1))=\theta$$
The measure $\mu^2$ is the product of $\mbox{\sl uniform}\{\pm 1\}$ and the measure on $[-1,1]$ whose density function is
$$w(x)=\begin{cases}
0 & |x|>\frac{1}{8}\\
\frac{8}{\pi\sqrt{1-\left( 8x\right)^2}} & |x|\le \frac{1}{8}
\end{cases}$$
Finally, $\mu=(1-\lambda) \mu^1+\lambda \mu^2$ for $\lambda >0$, which will be chosen later.

By lemma \ref{lem:C_is_small_main} (see remark \ref{rem:C_is_small_main}), there is a continuous normalized kernel $k'$ such that w.p. $\ge 1-2e\exp(-\frac{1}{\gamma})$ the function returned by the algorithm is of the form $f+b$ with $\|f\|_{H_{k'}}\le c\cdot m^3_A(\gamma)$ for some $c>0$ (depending only on $l$).
Let $e\in S^{d-1}$ be the vector from Lemma \ref{lemma:changes_slowly_high_dim}, corresponding to the kernel $k'$. The distribution $\mathcal D$ is the pullback of $\mu$ w.r.t. $e$.
By considering the affine functional $\Lambda_{e,0}$, it holds that $\Err_{\gamma}(\mathcal D)\le \lambda$. 

Let $g$ be the solution returned by the algorithm. With probability $\ge 1-\exp(-1/\gamma)$, $g=f+b$, where $f,b$ is a solution to program (\ref{eq:svm_equiv}) with $C=C_A(\gamma)$ and with an additive error $\le \sqrt{\gamma}$. Since the value of the zero solution for program (\ref{eq:svm_equiv}) is $l(0)$, it follows that
$$l(0)+\sqrt{\gamma}\ge \Err_{\mu,l}(g)=(1-\lambda)\Err_{\mu^1_e,l}(g)+\lambda\Err_{\mu^2_e,l}(g)$$
Thus, $\Err_{\mu^2_e,l}(g)\le\frac{l(0)+\sqrt{\gamma}}{\lambda}\le \frac{2l(0)}{\lambda}$. Combining Lemma \ref{lemma:L1_err}, Lemma \ref{lem:C_is_small_main}, and Lemma \ref{lemma:changes_slowly_high_dim} is follows that w.p. $\ge 1-(1+2e)\exp(-\frac{1}{\gamma})\ge 1-10\exp(-\frac{1}{\gamma})$, for $m=m_A(\gamma)$
\[
\left|\int_{\{x:\langle x,e\rangle =\gamma\}}g-\int_{\{x:\langle x,e\rangle =-\gamma\}}g\right|\le \frac{128l(0) \gamma K^{3.5}}{|\partial_+l(0)|\lambda} +10\cdot c\cdot K^{3.5}\cdot E\cdot m^3\cdot (r^K+s^d)
\]
By choosing $K=\Theta(\log (m))$, $\lambda=\Theta\left(\gamma K^{3.5}\right)=\Theta\left(\gamma \log^{3.5}(m)\right)$ and $d=\Theta(\log (m))$, we can make the last bound $\le  \frac{\alpha}{2}$. We claim that $\int_{\{x:\langle x,e\rangle =-\gamma\}}g>\frac{\alpha}{2}$. 
To see that, note that otherwise $\int_{\{x:\langle x,e\rangle =\gamma\}}g\le \alpha$ thus,
\begin{eqnarray*}
\mathbb E_{(x,y)\sim \mathcal D}l((f(x)+b)y) &=& \mathbb E_{(x,y)\sim \mathcal D}l(g(x)y)
\\
&\ge & \theta(1-\lambda)\cdot \int_{\{x:\langle x,e\rangle =\gamma \}}l(g(x))dx
\\
&\ge & \theta(1-\lambda)\cdot l\left(\int_{\{x:\langle x,e\rangle =\gamma \}}g(x)dx\right)
\\
&\ge & \theta\cdot l \left(\alpha\right)\cdot (1-\lambda)=\theta\cdot l \left(\alpha\right)+o(1)
\end{eqnarray*}
This contradict the optimality of $f,b$, as for $f'=0,b'=\beta$ it holds that
\begin{eqnarray*}
\mathbb E_{(x,y)\sim \mathcal D}l((f'(x)+b')y) &\le& \lambda l(-\beta)+\left(1-\lambda\right)\cdot\left(1-\theta)l(-\beta)+\theta\cdot l(\beta)\right)
\\
&= & (1-\theta)l(-\beta)+\theta\cdot l(\beta)+o(1)
\end{eqnarray*}

We can conclude now the proof of Theorem \ref{thm:main_1}. By choosing $d$ large enough and using Equation (\ref{eq:Levi}), we can
guarantee that $g|_{\{x:\langle x,e\rangle =-\gamma\}}$ is very concentrated around its expectation. In particular, if $(x,y)$ are sampled according to $\mathcal D$, then w.p. \mbox{$>0.5\cdot (1-\theta)\cdot (1-\lambda)=\Omega(1)$}, it holds that $yg(x)<0$. Thus, $\Err_{\mathcal D,0-1}(g)=\Omega(1)$, while $\Err_\gamma(\mathcal D)\le \lambda=O\left(\gamma \operatorname{poly}(\log(m))\right)$

To conclude the proof of Theorem \ref{thm:main_3}, we note that we can assume that $g$ is $\mathbb O(e)$-invariant. Otherwise, we can replace it with $\mathcal P_ef+b$. This does not increase $||f||_{H_k}$ nor $\Err_{\mathcal D,l}(f+b)$, thus, the solution $\mathcal P_ef+b$ is optimal as well. Now, it follows that $g|_{\{x:\langle x,e\rangle =-\gamma\}}$ is constant and we finish as before.
\proofbox

\subsubsection{Theorem \ref{thm:main_2}}
Let $L$ be the Lipschitz constant of $l$. Let $\beta>\alpha>0$ such that $l(\alpha)>l(\beta)$. Choose $0<\theta<1$ large enough so that $(1-\theta)l(-\beta)+\theta l(\beta)<\theta l(\alpha)$.
First, define probability measures $\mu^1,\mu^2,\mu^3$ and $\mu$ over $[-1,1]\times\{\pm 1\}$ as follows. 
$$\mu^1(\gamma,1)=\theta,\;\mu^1(-\gamma,-1)=1-\theta$$
$$\mu^2(-\gamma,1)=1$$
The measure $\mu^3$ is the product of $\mbox{\sl uniform} \{\pm 1\}$ and the measure over $[-1,1]$ whose density function is
$$w(x)=\begin{cases}
0 & |x|>\frac{1}{8}\\
\frac{8}{\pi\sqrt{1-\left( 8x\right)^2}} & |x|\le \frac{1}{8}
\end{cases}$$
Finally, $\mu=(1-\lambda_1-\lambda_2) \mu^1+\lambda_2 \mu^2+\lambda_3 \mu^3$ with $\lambda_2,\lambda_3 >0$ to be chosen later.

By lemma \ref{lem:C_is_small_main} (see remark \ref{rem:C_is_small_main}), there is a continuous normalized kernel $k'$ such that w.p. $\ge 1-2e\exp(-\frac{1}{\gamma})$ the function returned by the algorithm is of the form $f+b$ with $\|f\|_{H_{k'}}\le c\cdot m^3_A(\gamma)$ for some $c>0$ (depending only on $l$). Now, let $e\in S^{d-1}$ be the vector from Lemma \ref{lemma:changes_slowly_high_dim}, corresponding to the kernel $k'$. The distribution $\mathcal D$ is the pullback of $\mu$ w.r.t. $e$.
By considering the affine functional $\Lambda_{e,0}$, it holds that $\Err_{\gamma}(\mathcal D)\le \lambda_3+\lambda_2$.

Let $g$ be the solution returned by the algorithm. With probability $\ge 1-\exp(-1/\gamma)$, $g=f+b$, where $f,b$ is a solution to program (\ref{eq:svm_equiv}) with $C=C_A(\gamma)$ and with an additive error $\le \sqrt{\gamma}$. 
As in the proof of Theorem \ref{thm:main_1}, it holds that, w.p. $\ge 1-10\exp(-\frac{1}{\gamma})$ for $m=m_A(\gamma)$,
\begin{equation}\label{eq:7}
\left|\int_{\{x:\langle x,e\rangle =\gamma\}}g-\int_{\{x:\langle x,e\rangle =-\gamma\}}g\right|\le \frac{128l(0) \gamma K^{3.5}}{|\partial_+l(0)|\lambda_3} +10\cdot c\cdot K^{3.5} \cdot E\cdot m^3\cdot (r^K+s^d)
\end{equation}
Denote the last bound by $\epsilon$.
It holds that
\begin{equation}\label{eq:1}
\Err_{\mathcal D,l}(g) = (1-\lambda_2-\lambda_3)\mathbb E_{\mu^1_e}l(yg(x))
+ \lambda_2\mathbb E_{\mu^2_e}l(yg(x))
+ \lambda_3\mathbb E_{\mu^3_e}l(yg(x))
\end{equation}
Now, denote $\delta=\int_{\{x:\langle x,e\rangle =-\gamma\}}g$. It holds that
\begin{eqnarray}\label{eq:3}
\mathbb E_{\mu^1_e}l(yg(x)) &= & \theta\int_{\{x:\langle x,e\rangle =\gamma\}}l(g(x))+(1-\theta)\int_{\{x:\langle x,e\rangle =-\gamma\}}l(-g(x))\nonumber
\\
&\ge & \theta\cdot l\left( \int_{\{x:\langle x,e\rangle =\gamma\}}g\right)+(1-\theta)\cdot l\left(-\int_{\{x:\langle x,e\rangle =-\gamma\}}g\right)
\\
&\ge & \theta\cdot l(\delta)+(1-\theta)\cdot l(-\delta)-L\epsilon\nonumber
\end{eqnarray}
Thus,
\begin{eqnarray*}
\Err_{\mathcal D,l}(g) &\ge& (1-\lambda_2-\lambda_3)(\theta\cdot l(\delta)+(1-\theta)\cdot l(-\delta))-L\epsilon +\lambda_2\mathbb E_{\mu^2_e}l(yg(x))
\end{eqnarray*}
However, by considering the constant solution $\delta$, it follows that
\begin{eqnarray*}
\Err_{\mathcal D,l}(g) &\le & (1-\lambda_2-\lambda_3)(\theta l(\delta)+(1-\theta)\cdot l(-\delta)) +\lambda_2 \cdot l(\delta)+\lambda_3\frac{1}{2}\left(l(\delta)+l(-\delta)\right)+\sqrt{\gamma}
\\
&\le& (1-\lambda_2-\lambda_3)(\theta\cdot l(\delta)+(1-\theta)\cdot l(-\delta)) +\lambda_2\cdot l(\delta)+\lambda_3\cdot l(-|\delta|)+\sqrt{\gamma}
\end{eqnarray*}
Thus,
\begin{eqnarray}\label{eq:8}
\Err_{\mu^2_e,l}(g) &\le & \frac{L\epsilon}{\lambda_2}+l(\delta)+\frac{\lambda_3}{\lambda_2}l(-|\delta|)+\frac{\sqrt{\gamma}}{\lambda_2}
 \\
&= &\frac{L\cdot l(0) 128 \gamma K^{3.5}}{|\partial_+l(0)|\lambda_2\lambda_3}+\frac{10\cdot c \cdot L \cdot K^{3.5}}{\lambda_2}\cdot E\cdot m^3\cdot (r^K+s^d)+l(\delta)+\frac{\lambda_3}{\lambda_2}l(-|\delta|)+\frac{\sqrt{\gamma}}{\lambda_2}\nonumber
\end{eqnarray}
Now, relying on the assumption that $\gamma\cdot\log^8(m)=o(1)$, it is possible to choose $\lambda_2=\Theta\left(\sqrt{\gamma} K^{4}\right)=\Theta\left(\sqrt{\gamma} \log^{4}(m)\right)$, $\lambda_3=\sqrt{\gamma}$, $K=\Theta(\log (m/\gamma))$,  and $d=\Theta(\log (m/\gamma))$ such that 
the bound in Equation (\ref{eq:7}), $\frac{L\cdot l(0)128 \gamma K^{3.5}}{|\partial_+l(0)|\lambda_2\lambda_3} +\frac{10\cdot c\cdot  K^{3.5}}{\lambda_2}\cdot E\cdot m^3\cdot (r^K+s^d)$, $\lambda_2$, $\lambda_3$ and $\frac{\lambda_3}{\lambda_2}$ are all $o(1)$.

Since the bound in Equation (\ref{eq:7}) is $o(1)$, it follows, as in the proof of Theorem \ref{thm:main_1}, that $l(\delta)\le l\left(\frac{\alpha}{2}\right)$ and consequently, $0<\frac{\alpha}{2}\le \delta$.
From equations (\ref{eq:1}) and (\ref{eq:3}), it follows that
$$l(-|\delta|)=l(-\delta)
\le \frac{L\epsilon+\frac{\Err_{\mathcal D,l}(g)}{1-\lambda_2-\lambda_3}}{1-\theta}
\le \frac{L\epsilon+\frac{2l(0)}{1-\lambda_2-\lambda_3}}{1-\theta}=O(1)$$
It now follows from Equation (\ref{eq:8}) that
$$\mathbb E_{(x,y)\sim\mu_2}l(g(x)y)=\Err_{\mu^2_e,l}(g) \le l\left(\frac{\alpha}{2}\right)+o(1)$$
By Markov's inequality, 
$$\Pr_{(x,y)\sim\mu_2}\left(l(g(x)y)\ge l(0)\right)\le\frac{l\left(\frac{\alpha}{2}\right)+o(1)}{l(0)}$$
Thus, if $(x,y)$ are chosen according to $\mu^2_e$, then w.p. $>\frac{l(0)-l\left(\frac{\alpha}{2}\right)}{l(0)}-o(1)$, $l(g(x))< l(0)\Rightarrow g(x)>0$. Since the marginal distributions of $\mu^1_e$ and $\mu^2_e$ are the same, it follows that, if $(x,y)$ are chosen according to $\mathcal D$, then w.p. $>\left(\frac{l(0)-l\left(\frac{\alpha}{2}\right)}{l(0)}-o(1)\right)\cdot(1-\lambda_2-\lambda_3)\cdot(1-\theta)=\Omega(1)$, $yg(x)<0$. Thus, $\Err_{\mathcal D,0-1}(g)=\Omega(1)$ while $\Err_\gamma(\mathcal D)\le \lambda_2+\lambda_3=O\left(\sqrt{\gamma}\poly(\log(m))\right)$.
\proofbox

\subsubsection{The integrality gap -- Theorem \ref{thm:main_4}}
Our first step is a reduction to the hinge loss. Let $a=\partial_+l(0)$. Define
$$l^*(x)=\begin{cases} ax+1 & x \le \frac{1}{-a}\\ 0 & o/w\end{cases}$$
it is not hard to see that $l^{*}$ is a convex surrogate satisfying $\forall x,\;l^{*}(x)\le l(x)$ and $\partial_+l^*(0)=\partial_+l(0)$. Thus, if we substitute $l$ with $l^{*}$, we just decrease the integrality gap, hence can assume that $l=l^*$. Now, we note that if we consider program (\ref{eq:svm_equiv}) with $l=l^*$ the inegrality gap of coincides with what we get by replacing $C$ with $|a|\cdot C$ and $l^*$ with the hinge loss. To see that, note that for every $f\in H_k,b\in\reals$, $\Err_{\mathcal D,l^*}(f+b)=\Err_{\mathcal D,\hinge}(|a|\cdot f+|a|\cdot b)$, thus, minimizing $\Err_{\mathcal D,l^*}$ over all functions $f\in H_k$ that satisfy $||f||_{H_k}\le C$ is equivalent to minimizing $\Err_{\mathcal D,\hinge}$ over all functions $f\in H_k$ that satisfy $||f||_{H_k}\le |a|\cdot C$. Thus, it is enough to prove the Theorem for $l=l_{\hinge}$.

Next, we show that we can assume that the embedding is symmetric (i.e., correspond to a symmetric kernel). As the integrality gap is at least as large as the approximation ratio, using Theorem \ref{thm:main_3} this will complete our argument. (The reduction to the hinge loss yields bounds with universal constants in the asymptotic terms).

Let $\gamma>0$ and let $\mathcal D$ be a distribution on $S^{d-1}\times \{\pm 1\}$. It is enough to find (a possibly different) distribution $\mathcal D_1$ with the same $\gamma$-margin error as $\mathcal D$, for which the optimum of program (\ref{eq:svm_equiv}) (with $l=l_{\hinge}$) is not smaller than the optimum of the program
\begin{eqnarray}\label{eq:svm_1_hinge_sym_2}
\min &&  \Err_{\mathcal D,\hinge}\left(f+b\right) \nonumber \\
\text{s.t.} && f\in H_{k_s},\;b\in\reals\\
&& ||f||_{H_{k_s}}\le C\nonumber
\end{eqnarray}
Denote the optimal value of program (\ref{eq:svm_1_hinge_sym_2}) by $\alpha$ and assume, towards contradiction, that whenever $\Err_\gamma(\mathcal D_1)=\Err_\gamma(\mathcal D)$, the optimum of program (\ref{eq:svm_equiv}) is strictly less then $\alpha$. 

For every $A\in \mathbb O(d)$, let $\mathcal D_A$, be the distribution of the r.v. $(Ax,y)\in S^{d-1}\times\{\pm 1\}$, where $(x,y)\sim\mathcal D$. Since clearly $\Err_\gamma(\mathcal D_A)=\Err_\gamma(\mathcal D)$, there exist $f_A\in H_k$ and $b_A \in\reals$ such that $||f_A||_{H_k}\le C$ and $\Err_{\mathcal D_A,\hinge}(g_A)<\alpha$, where $g_A:=f_A+b_A$.
Define $f\in H_{k_s}$ by $f(x)=\int_{\mathbb O(d)}f_A(Ax)dA$ and let 
$b=\int_{\mathbb O(d)}b_AdA$ and $g=f+b$. By Theorem \ref{thm:symmetrization}, $||f||_{H_{k_s}}\le C$. Finally, for $l=l_{\hinge}$,
\begin{eqnarray*}
\Err_{\mathcal D,\hinge}(g) &=& \mathbb E_{(x,y)\sim \mathcal D}l(yg(x))
\\
&=& \mathbb E_{(x,y)\sim \mathcal D}l(y\mathbb E_{A\sim \mathbb O(d)}g_A(Ax))
\\
&\le & \mathbb E_{(x,y)\sim \mathcal D}\mathbb E_{A\sim \mathbb O(d)}l(yg_A(Ax))
\\
&= & \mathbb E_{A\sim \mathbb O(d)}\mathbb E_{(x,y)\sim \mathcal D}l(yg_A(Ax))
\\
&= & \mathbb E_{A\sim \mathbb O(d)}\mathbb E_{(x,y)\sim \mathcal D_A}l(yg_A(x)) < \alpha
\end{eqnarray*}
Contrary to the assumption that $\alpha$ is the optimum of program (\ref{eq:svm_1_hinge_sym_2}).

\subsubsection{Finite dimension - Theorems \ref{thm:main_1_fin_dim} and \ref{thm:main_4_fin_dim}}
Let $V\subseteq C(S^{d-1})$ be the linear space 
$\{\Lambda_{v,b}\circ\psi:v\in\reals^m,b\in\reals\}$ and denote $\bar W=\{\Lambda_{v,b}\circ\psi:v\in W,b\in\reals\}$.
We note that $\dim(V)\le m+1$. Instead of program (\ref{eq:svm_finite_dim}) we consider the equivalent formulation
\begin{eqnarray}\label{eq:svm_finite_dimension}
\min &&  \Err_{\mathcal D,l}\left(f\right) \nonumber \\
\text{s.t.} && f\in \bar W
\end{eqnarray}
The following lemma is very similar to lemma \ref{lem:fin_dim_to_inner_prod}, but with better dependency on $m$ ($m^{1.5}$ instead of $m^2$).

\begin{lemma}\label{lem:reduction_finite_dim_to_kernel}
Let $l$ be a convex surrogate and let $V\subset C(S^{d-1})$ an $m$-dimensional vector space. There exists a continuous kernel $k:S^{d-1}\times S^{d-1}\to\reals$ with $\sup_{x\in S^{d-1}}k(x,x)\le 1$ such that $H_k=V$ as a vector space and there exists a probability measure $\mu_N$ such that
$$\forall f\in V,\; ||f||_{H_k}\le \frac{2m^{1.5}}{|\partial_+l(0)|}\Err_{\mu_N,l}(f)$$ 
\end{lemma}
\proof
Let $\psi:S^{d-1}\to V^*$ be the evaluation operator. It maps each $x\in S^{d-1}$ to the linear functional $f\in V\mapsto f(x)$.  
We claim that
\begin{enumerate}
\item\label{fin_dim_item_1}
$\psi$ is continuous,
\item\label{fin_dim_item_2}
$\mbox{aff}(\psi(S^{d-1})\cup-\psi(S^{d-1}))=V^*$,
\item\label{fin_dim_item_3}
$V=\{v^{**}\circ \psi:v^{**}\in V^{**}\}$.
\end{enumerate}
Proof of~\ref{fin_dim_item_1}: We need to show that $\psi(x_n)\to\psi(x)$ if $x_n\to x$. Since $V^*$ is finite dimensional, it suffices to show that $\psi(x_n)(f)\to\psi(x)(f)$ for every $f\in V$, which follows from the continuity of $f$.\\
Proof of~\ref{fin_dim_item_2}: Note that $0\in U=\mbox{aff}(\psi(S^{d-1})\cup-\psi(S^{d-1}))$, so $U$ is a linear space. Now, define $T:U^*\to V$ via $T(u^*)=u^*\circ\psi$. We claim that $T$ is onto, whence $\dim(U)=\dim(U^*)=\dim(V)=\dim(V^*)$, so that $U=V^*$. 
Indeed, for $f\in V$, let $u^*_f\in U^*$ be the functional $u^*_f(u)=u(f)$. Now, $T(u^*_f)(x)=u^*_f(\psi(x))=\psi(x)(f)=f(x)$, thus $T(u^*_f)=f$.\\
Proof of~\ref{fin_dim_item_3}: From $U=V^*$ it follows that $U^*=V^{**}$, so that the mapping $T:V^{**}\to V$ is onto, showing that $V=\{v^{**}\circ \psi:v^{**}\in V^{**}\}$.

Let us apply John's Lemma
to $K=\operatorname{conv}(\psi(S^{d-1})\cup-\psi(S^{d-1}))$. It yields an inner product on $V^*$ with $K$ contained in the unit ball and containing the ball around $0$ with radius $\frac{1}{\sqrt{m}}$. Let $k$ be the kernel $k(x,y)=\langle\psi(x),\psi(y)\rangle$. Since $\psi$ is continuous, $k$ is continuous as well. By Theorem \ref{thm:RKHS_embedding} and since $T$ is onto, it follows that, as a vector space, $V=H_k$. Since $K$ is contained in the unit ball, it follows that $\sup_{x\in S^{d-1}}k(x,x)\le 1$. It remains to prove the existence of the measure $\mu_N$.
 
Let $e_1,\ldots,e_m\in V^*$ be an orthonormal basis. For every $i\in [m]$, choose $(x_i^1,y_i),\ldots,(x_i^{m+1},y_i)\in S^{d-1}\times \{\pm 1\}$ and $\lambda_i^1,\ldots,\lambda_i^{m+1}\ge 0$ such that $\sum_{j=1}^{m+1}\lambda_i^j=1$ and $\frac{1}{\sqrt{m}}e_i=\sum_{j=1}^{m+1}\lambda_i^jy_i\psi(x^j_i)$. Define $\mu_N(x^j_i,1)=\mu_N(x^j_i,-1)=\frac{\lambda_i^j}{2m}$.

Let $f\in V$. By Theorem \ref{thm:RKHS_embedding} there exists $v\in V^*$ such that $f=\Lambda_{v,0}\circ \psi$ and $||f||_{H_k}=||v||_{V^*}$. It follows that, for $a=\partial_+l(0)$, 

\begin{eqnarray*}
\Err_{\mu_N,l}(f) &=& \sum_{i=1}^m\sum_{j=1}^{m+1}\frac{\lambda_i^j}{2m}\left[l(y_if(x_i^j))+l(-y_if(x_i^j))\right]
\\
&\ge & \frac{1}{2m}\sum_{i=1}^m\left[l\left(\sum_{j=1}^{m+1}\lambda_i^jy_if(x_i^j)\right)
+l\left(-\sum_{j=1}^{m+1}\lambda_i^jy_if(x_i^j)\right)\right]
\\
&= & \frac{1}{2m}\sum_{i=1}^m\left[l\left(\sum_{j=1}^{m+1}\lambda_i^jy_i\langle v,\psi(x_i^j)\rangle\right)+l\left(-\sum_{j=1}^{m+1}\lambda_i^jy_i\langle v,\psi(x_i^j)\rangle\right)\right]
\\
&= & \frac{1}{2m}\sum_{i=1}^ml\left(\langle v,\sum_{j=1}^{m+1}\lambda_i^jy_i\psi(x_i^j)\rangle\right)+
l\left(-\langle v,\sum_{j=1}^{m+1}\lambda_i^jy_i\psi(x_i^j)\rangle\right)
\\
&= & \frac{1}{2m}\sum_{i=1}^ml\left(\langle v,\frac{e_i}{\sqrt{m}}\rangle\right)+l\left(-\langle v,\frac{e_i}{\sqrt{m}}\rangle\right)
\\
&\ge & \frac{1}{2m}\sum_{i=1}^ml\left(-\frac{|\langle v,e_i\rangle|}{\sqrt{m}}\right)
\\
&\ge & \frac{|a|}{2m^{1.5}}\sum_{i=1}^m|\langle v,e_i\rangle|
\\
&\ge & \frac{|a|}{2m^{1.5}}||v||_{V^*}=\frac{|a|}{2m^{1.5}}||f||_{H_k}
\end{eqnarray*}
\proofbox

\proof (of Theorem \ref{thm:main_1_fin_dim})
Let $L$ be the Lipschitz constant of $l$. Let $\beta>\alpha>0$ such that $l(\alpha)>l(\beta)$. Choose $0<\theta<1$ large enough so that $(1-\theta)l(-\beta)+\theta l(\beta)<\theta l(\alpha)$.
First, define probability measures $\mu^1,\mu^2,\mu^3$ and $\mu$ over $[-1,1]\times\{\pm 1\}$ as follows. 
$$\mu^1(\gamma,1)=\theta,\;\mu^1(-\gamma,-1)=1-\theta$$
$$\mu^2(-\gamma,1)=1$$
The measure $\mu^3$ is the product of $\mbox{\sl uniform}\{\pm 1\}$ and the measure over $[-1,1]$ whose density function is
$$w(x)=\begin{cases}
0 & |x|>\frac{1}{8}\\
\frac{8}{\pi\sqrt{1-\left( 8x\right)^2}} & |x|\le \frac{1}{8}
\end{cases}$$
Let $k$, $\mu_N$ be the distribution and kernel from Lemma \ref{lem:reduction_finite_dim_to_kernel}.
Now, let $e\in S^{d-1}$ be the vector from Lemma \ref{lemma:changes_slowly_high_dim}. 
We define the distribution $\mathcal D$ corresponding to the measure
$$\mu=(1-\lambda_2-\lambda_3-\lambda_N)\mu_e^1+\lambda_2\mu^2_e
+\lambda_3\mu^3_e+\lambda_N\mu_N$$
By considering the affine functional $\Lambda_{e,0}$, it holds that $\Err_{\gamma}(\mathcal D)\le \lambda_3+\lambda_2+\lambda_N$.

Let $g$ be the solution returned by the algorithm. With probability $\ge 1-\exp(-1/\gamma)$, $g=f+b$, where $f,b$ is a solution to program (\ref{eq:svm_finite_dimension}) with an additive error $\le \sqrt{\gamma}$.

Denote $||g||_{H_k}=C$. By Lemma \ref{lem:reduction_finite_dim_to_kernel}, it holds that
\begin{eqnarray*}
C &\le & \frac{2m^{1.5}}{|\partial_+l(0)|}\Err_{\mu_N,l}(g)
\\
&\le & \frac{2m^{1.5}}{|\partial_+l(0)|}\frac{\Err_{\mu,l}(g)}{\lambda_N}
\\
&\le & \frac{2m^{1.5}}{|\partial_+l(0)|}\frac{l(0)}{\lambda_N}
\end{eqnarray*}
As in the proof of Theorem \ref{thm:main_1}, it holds that
\begin{equation}\label{eq:9}
\left|\int_{\{x:\langle x,e\rangle =\gamma\}}g-\int_{\{x:\langle x,e\rangle =-\gamma\}}g\right|\le \frac{128 l(0)\gamma K^{3.5}}{|\partial_+l(0)|\lambda_3} +10\cdot K^{3.5}\cdot E\cdot C\cdot (r^K+s^d)
\end{equation}
Denote the last bound by $\epsilon$.
It holds that
\begin{equation}\label{eq:10}
\Err_{\mathcal D,l}(g) = (1-\lambda_2-\lambda_3-\lambda_N)\mathbb E_{\mu^1_e}l(yg(x))
+ \lambda_2\mathbb E_{\mu^2_e}l(yg(x))
+ \lambda_3\mathbb E_{\mu^3_e}l(yg(x))
+ \lambda_N\mathbb E_{\mu_N}l(yg(x))
\end{equation}
Now, denote $\delta=\int_{\{x:\langle x,e\rangle =-\gamma\}}g$. It holds that
\begin{eqnarray}\label{eq:11}
\mathbb E_{\mu^1_e}l(yg(x)) &= & \theta\int_{\{x:\langle x,e\rangle =\gamma\}}l(g(x))+(1-\theta)\int_{\{x:\langle x,e\rangle =-\gamma\}}l(-g(x))\nonumber
\\
&\ge & \theta\cdot l\left( \int_{\{x:\langle x,e\rangle =\gamma\}}g\right)+(1-\theta)\cdot l\left(-\int_{\{x:\langle x,e\rangle =-\gamma\}}g\right)
\\
&\ge & \theta\cdot l(\delta)+(1-\theta)\cdot l(-\delta)-L\epsilon\nonumber
\end{eqnarray}
Thus,
\begin{eqnarray*}
\Err_{\mathcal D,l}(g) &\ge& (1-\lambda_2-\lambda_3-\lambda_N)(\theta\cdot l(\delta)+(1-\theta)\cdot l(-\delta))-L\epsilon +\lambda_2\mathbb E_{\mu^2_e}l(yg(x))
\end{eqnarray*}
However, by considering the constant solution $\delta$, it follows that
\begin{eqnarray*}
\Err_{\mathcal D,l}(g) &\le & (1-\lambda_2-\lambda_3-\lambda_N)(a\cdot l(\delta)+(1-\theta)\cdot l(-\delta)) +\lambda_2 \cdot l(\delta)+(\lambda_3+\lambda_N)\frac{1}{2}\left(l(\delta)+l(-\delta)\right)+\sqrt{\gamma}
\\
&\le& (1-\lambda_2-\lambda_3-\lambda_N)(\theta\cdot l(\delta)+(1-\theta)\cdot l(-\delta)) +\lambda_2\cdot l(\delta)+(\lambda_3+\lambda_N)\cdot l(-|\delta|)+\sqrt{\gamma}
\end{eqnarray*}
Thus,
\begin{eqnarray}\label{eq:12}
\Err_{\mu^2_e, l}(g) &\le & \frac{L\epsilon}{\lambda_2}+l(\delta)+\frac{\lambda_3+\lambda_N}{\lambda_2}l(-|\delta|)+\frac{\sqrt{\gamma}}{\lambda_2}
\\
&\le &\frac{L\cdot l(0) 128 \gamma K^{3.5}}{|\partial_+l(0)|\lambda_2\lambda_3} +\frac{10\cdot L\cdot K^{3.5}}{\lambda_2}\cdot E\cdot C\cdot (r^K+s^d)+l(\delta)+\frac{\lambda_3+\lambda_N+\sqrt{\gamma}}{\lambda_2}l(-|\delta|)\nonumber
\end{eqnarray}
Now, relying on the assumption that $\gamma\cdot\log^8(C)=o(1)$, it is possible to choose $\lambda_2=\Theta\left(\sqrt{\gamma} K^{4}\right)=\Theta\left(\sqrt{\gamma} \log^{4}(C)\right)$, $\lambda_3=\sqrt{\gamma}$, $K=\Theta(\log (C/\gamma))$, $\lambda_N=\gamma$  and $d=\Theta(\log (C/\gamma))$ such that 
the bound in Equation (\ref{eq:9}), $\frac{L\cdot l(0)128 \gamma K^{3.5}}{|\partial_+l(0)|\lambda_2\lambda_3} +\frac{10 K^{3.5}}{\lambda_2}\cdot E\cdot C\cdot (r^K+s^d)$, $\lambda_2$, $\lambda_3$, $\lambda_N$ and $\frac{\lambda_3+\lambda_N+\sqrt{\gamma}}{\lambda_2}$ are all $o(1)$.

Since the bound in Equation (\ref{eq:9}) is $o(1)$, it follows, as in the proof of Theorem \ref{thm:main_1}, that $l(\delta)\le l\left(\frac{\alpha}{2}\right)$ and consequently, $0<\frac{\alpha}{2}\le \delta$.
From equations (\ref{eq:10}) and (\ref{eq:11}), it follows that
$$l(-|\delta|)=l(-\delta)
\le \frac{L\epsilon+\frac{\Err_{\mathcal D,l}(g)}{1-\lambda_2-\lambda_3-\lambda_N}}{1-\theta}
\le \frac{L\epsilon+\frac{2l(0)}{1-\lambda_2-\lambda_3-\lambda_N}}{1-\theta}=O(1)$$
It now follows from Equation (\ref{eq:12}) that
$$\mathbb E_{(x,y)\sim\mu_2}l(g(x)y)=\Err_{\mu^2_e,l}(g) \le l\left(\frac{\alpha}{2}\right)+o(1)$$
By Markov's inequality, 
$$\Pr_{(x,y)\sim\mu_2}\left(l(g(x)y)\ge l(0)\right)\le\frac{l\left(\frac{\alpha}{2}\right)+o(1)}{l(0)}$$
Thus, if $(x,y)$ are chosen according to $\mu^2_e$, then w.p. $>\frac{l(0)-l\left(\frac{\alpha}{2}\right)}{l(0)}-o(1)$, $l(g(x))< l(0)\Rightarrow g(x)>0$. Since the marginal distributions of $\mu^1_e$ and $\mu^2_e$ are the same, it follows that, if $(x,y)$ are chosen according to $\mathcal D$, then w.p. $>\left(\frac{l(0)-l\left(\frac{\alpha}{2}\right)}{l(0)}-o(1)\right)\cdot(1-\lambda_2-\lambda_3-\lambda_N)\cdot(1-\theta)=\Omega(1)$, $yg(x)<0$. Thus, $\Err_{\mathcal D,0-1}(g)=\Omega(1)$ while $\Err_\gamma(\mathcal D)\le \lambda_2+\lambda_3+\lambda_N=O\left(\sqrt{\gamma}
\poly(\log(C))\right)
=O\left(\sqrt{\gamma}\poly(\log(m/\gamma))\right)$.
\proofbox

\proof (of Theorem \ref{thm:main_4_fin_dim})
As in the proof of Theorem \ref{thm:main_4}, we can assume w.l.o.g. that $l=l_{\hinge}$. 
Let $k,\mu_N$ be the measure and the kernel from Lemma \ref{lem:reduction_finite_dim_to_kernel}. Let $C=2m^{1.5}/\gamma$. By (the proof of) Theorem \ref{thm:main_4}, there exists a probability measure $\bar \mu$ over $S^{d-1}\times\{\pm 1\}$ such that for every $f\in H_k$ with $||f||_{H_k}\le C$ it holds that $\Err_{\bar \mu,l}(f)=\Omega(1)$ but $\Err_{\gamma}(\bar\mu)=O(\gamma\cdot\poly(\log(C)))$. Consider the distribution $\mu=(1-\gamma)\bar \mu+\gamma\mu_N$. It still holds that $\Err_{\gamma}(\bar\mu)=O(\gamma\cdot\poly(\log(C)))=O(\gamma\cdot\poly(\log(m/\gamma)))$. Let $f$ be an optimal for program (\ref{eq:svm_finite_dimension}). We have that
$1\ge \Err_{\mu,l}(f)\ge \gamma\cdot \Err_{\mu_N,l}(f)$. By Lemma \ref{lem:reduction_finite_dim_to_kernel}, $||f||_{H_k}\le C$. Thus, $\Err_{\mu,l}(f)\ge (1-\gamma)\Err_{\bar\mu,l}(f)=\Omega(1)$.   
\proofbox

\section{Choosing a surrogate according to the margin}\label{sec:choosing_l}
The purpose of this section is to demonstrate the subtleties relating to the possibility of choosing a convex surrogate $l$ according to the margin $\gamma$. 
Let $k:B\times B\to\reals$ be the kernel
$$k(x,y)=\frac{1}{1-\frac{1}{2}\inner{x,y}_H}$$
and let $\psi:B\to H_1$ be a corresponding embedding (i.e., $k(x,y)=\inner{\psi(x),\psi(y)}_{H_1}$).
In \citep{ShalevShSr11} it has been shown that the solution $f,b$ to Program (\ref{eq:svm_1_hinge}), with $C=C(\gamma)=\poly(\exp(1/\gamma\cdot\log(1/\gamma)))$ and the embedding $\psi$, satisfies
$$\Err_{\hinge}(f+b)\le \Err_{\gamma}(\cd)+\gamma ~.$$
Consequently, every approximated solution to the Program with an additive error of at most $\gamma$ will have a 0-1 loss bounded by $\Err_\gamma(\cd)+2\gamma$.

For every $\gamma$, define a $1$-Lipschitz convex surrogate by
$$l_\gamma(x)=\begin{cases}
1-x & x\le 1/C(\gamma)\\
1-1/C(\gamma) & x\ge 1/C(\gamma)
\end{cases}$$
\begin{claim}\label{claim:choosing_l}
A function $g:B\to\reals$ is a solutions to Program (\ref{eq:svm_1_general}) with $l=l_\gamma$, $C=1$ and the embedding $\psi$, if and only if $C(\gamma)\cdot g$ is a solutions to Program (\ref{eq:svm_1_hinge}) with $C=C(\gamma)$ and the embedding $\psi$.
\end{claim}
We postpone the proof to the end of the section. We note that Program
(\ref{eq:svm_1_general}) with $l=l_\gamma$, $C=1$ and the embedding
$\psi$, have a complexity of $1$, according to our
conventions. Moreover, by Claim \ref{claim:choosing_l}, the optimal
solution to it has a 0-1 error of at most
$\Err_\gamma(\cd)+\gamma$. Thus, if $A$ is an algorithm that is only
obligated to return an approximated solution to Program
(\ref{eq:svm_1_general}) with $l=l_\gamma$, $C=1$ and the embedding
$\psi$, we cannot lower bound its approximation ratio. In particular,
our Theorems regarding the approximation ratio are no longer true, as
currently stated, if the algorithms are allowed to choose the
surrogate according to $\gamma$. One might be tempted to think that by
the above construction (i.e. taking $\psi$ as our embedding, choosing
$C=1$ and $l=l_\gamma$, and approximate the program upon a sample of
size $\poly(1/\gamma)$), we have actually gave $1$-approximation
algorithm. The crux of the matter is that algorithms that approximate
the program according to a finite sample of size $\poly(1/\gamma)$ are
only guaranteed to find a solution with an additive error of
$\poly(\gamma)$. For the loss $l_\gamma$, such an additive error is meaningless: Since for every function $f$, $\Err_{\cd,l_\gamma}(f)\ge 1-1/C(\gamma)$, the $0$ solution has an additive error of $\poly(\gamma)$. Therefore, we cannot argue that the solution returned by the algorithm will have a small 0-1 error. Indeed we anticipate that the algorithm we have described will suffer from serious over-fitting.

To summarize, we note that the lower bounds we have proved, relies on the fact that the optimal solutions of the programs we considered are very bad. For the algorithm we sketched above, the optimal solution is very good. However, guaranties on approximated solutions obtained from a polynomial sample are meaningless.  We conclude that lower bounds for such algorithms will have to involve over-fitting arguments, which are out of the scope of the paper.

\proof (of claim \ref{claim:choosing_l})
Define
$$l^*_\gamma(x)=\begin{cases}
1-C(\gamma)x & x\le \frac{1}{C(\gamma)}\\
0 & x\ge \frac{1}{C(\gamma)}
\end{cases}$$
Since $l^*_\gamma(x)=C(\gamma)\cdot(l_\gamma(x)-(1-\frac{1}{C(\gamma)}))$, it follows that the solutions to Program (\ref{eq:svm_1_general}) with $l=l^*_\gamma$, $C=1$ and $\psi$ coincide with the solutions with $l=l_\gamma$, $C=1$ and $\psi$. Now, we note that, for every function $f:B\to\reals$,
$$\Err_{\cd,l_\gamma^*}(f)=\Err_{\cd,\hinge}(C(\gamma)\cdot f)$$
Thus, $w,b$ minimizes $\Err_{\cd,l_\gamma^*}(\Lambda_{w,b}\circ\psi)$ under the restriction that $\|w\|\le 1$ if and only if $C(\gamma)\cdot w,C(\gamma)\cdot b$ minimizes $\Err_{\cd,\hinge}(\Lambda_{w,b}\circ\psi)$ under the restriction that $\|w\|\le C(\gamma)$.
\proofbox
\paragraph{Acknowledgements:}
Amit Daniely is a recipient of the Google Europe Fellowship in Learning Theory, and this research is supported in part by this Google Fellowship. Nati Linial is supported by grants from ISF, BSF and I-Core. Shai Shalev-Shwartz is supported by the Israeli Science Foundation grant number 590-10. 
\bibliography{bib}

\end{document}